\documentclass{article}

\usepackage[preprint]{neurips_2022}

\usepackage{natbib}
\setcitestyle{numbers,square}

\usepackage[utf8]{inputenc} 
\usepackage[T1]{fontenc}    
\usepackage{url}            
\usepackage{booktabs}       
\usepackage{amsfonts}       
\usepackage{nicefrac}       
\usepackage{microtype}      
\usepackage{xcolor}         
\usepackage{amsmath}
\usepackage{graphicx}
\usepackage{multirow}
\usepackage{booktabs}
\usepackage{caption}
\usepackage{subcaption}
\usepackage{float}

\usepackage[pagebackref,breaklinks,colorlinks]{hyperref}

\usepackage[capitalize]{cleveref}
\crefname{section}{Sec.}{Secs.}
\Crefname{section}{Section}{Sections}
\Crefname{table}{Table}{Tables}
\crefname{table}{Tab.}{Tabs.}

\title{Cascade Luminance and Chrominance for Image Retouching: More Like Artist}

\author{
Hailong Ma\\
Tsinghua University\\Huawei Technologies
\And
Sibo Feng\\
Huawei Technologies\\
\And
Xi Xiao\\
Tsinghua University
\AND
Chenyu Dong\\
Tsinghua University\\Huawei Technologies
\And
Xingyue Cheng\\
Tsinghua University}
\begin{document}

\maketitle

\begin{abstract}
Photo retouching aims to adjust the luminance, contrast, and saturation of the image to make it more human aesthetically desirable. However, artists' actions in photo retouching are difficult to quantitatively analyze. By investigating their retouching behaviors, we propose a two-stage network that brightens images first and then enriches them in the chrominance plane. Six pieces of useful information from image EXIF are picked as the network's condition input. Additionally, hue palette loss is added to make the image more vibrant. Based on the above three aspects, Luminance-Chrominance Cascading Net(LCCNet) makes the machine learning problem of mimicking artists in photo retouching more reasonable. Experiments show that our method is effective on the benchmark MIT-Adobe FiveK dataset, and achieves state-of-the-art performance for both quantitative and qualitative evaluation. 
\end{abstract}


\section{Introduction}\label{sec. introduction}
Raw images, or raw snapshots that have not been aesthetically improved by artistic intent, can significantly boost visual appeal and expressiveness with the help of computer interactive color adjustments' editing software such as Photoshop~\cite{tomas1987photoshop} and Lightroom~\cite{shadow2002lightroom}. However, the proficient use of these tools requires years of experience and the retouching procedures become tedious and time-consuming as images grow in quantity. Automating image retouching task aims to simulate the artists' operation to automatically beautify the image, which is the main objective of this research.

Traditional automatic retouching methods use equilibrium operations to decorate pictures, e.g. gray-world~\cite{finlayson2004Shades}, histogram equalization~\cite{naik2003hue} and its many variants~\cite{stark2000histogram,duan2004novalhist,arici2009histmodification,agaian2007transform,xu2014equalization}. Alternatively, filters such as fast bilateral filtering~\cite{durand2002Fast} and local Laplacian operator~\cite{aubry2014Fast} can be used to enhance the visual quality of images. These methods rely on prior knowledge to design fixed features, which is difficult to deal with rich scenes. Bychkovsky et al.~\cite{vladimir2011fivek} constructed a large image retouching dataset MIT-Adobe FiveK, which is widely used by image retouching tasks. It contains 5000 original images with corresponding five experts' retouching results, and the intermediate operations for retouching are recorded. 

\begin{figure}
\centering
\includegraphics[width=\textwidth]{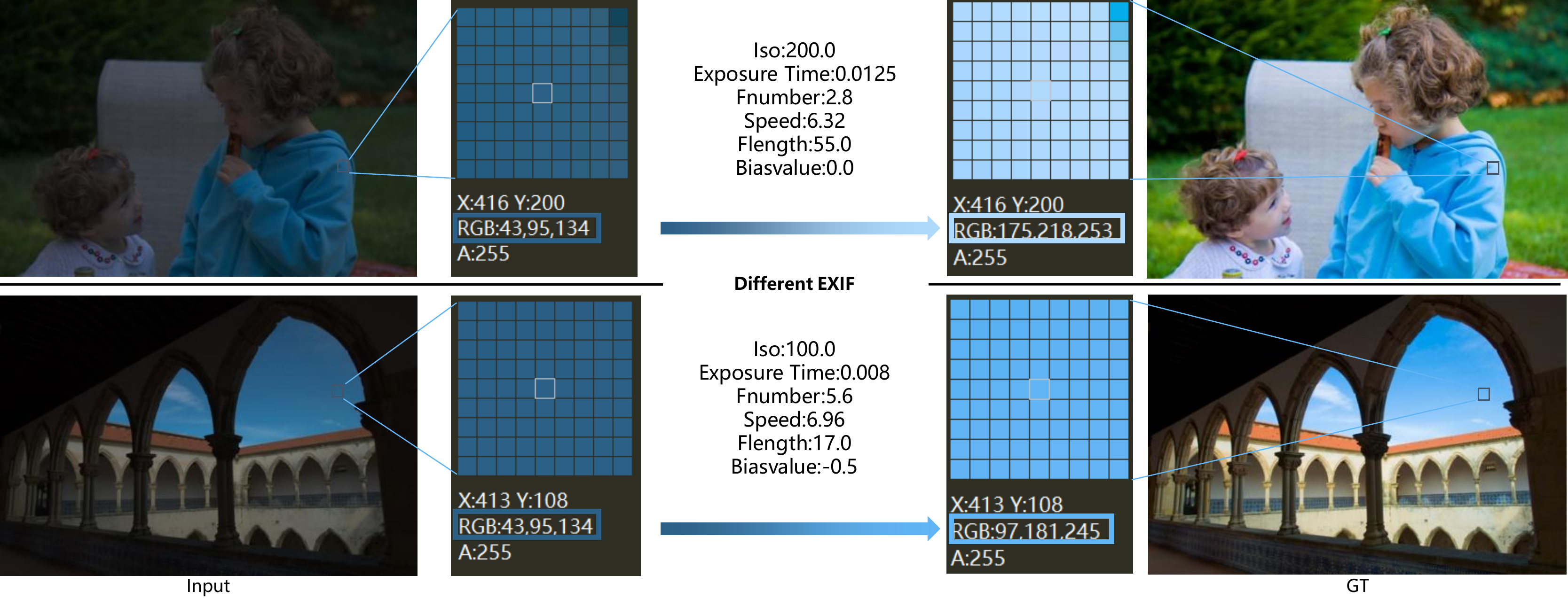}
\caption{One-to-many color mapping and their corresponding EXIF}
\label{fig:exifmapping}
\end{figure}

Deep networks use a self-learning feature expression method and can better deal with rich scenes in automatic image retouching. A deep learning-based method was proposed by Yan et.al in 2016~\cite{yan2016automatic}, since then, the deep neural network-based approaches have been more adopted. Roughly speaking, they can be divided into two types: global-based and local-based. The former algorithms focus on overall adjustments, these methods can be divided into curve-based~\cite{Song2021starenhancer,guo2020Zero}, table based~\cite{zeng20203Dlut,yang2022AdaInt,wang2021Real}, and other global coefficient adjustment methods~\cite{he2020csrnet,park2018distort,zhang2021STAR,zhao2021recurrent,shi2021guided,liang2021LPTN}. Local-based methods~\cite{gharbi2017hdrnet,isola2017P2P,wang2019DUPE,pan2021MIEGAN,kim2021representative,yu2018deepexposure,kim2020globallocal,moran2020deeplpf} adjust the image at the pixel level. All these methods use only images as input, but a single image can only provide limited scene information since the shooting process is a dimensionality reduction process. In reality, as seen in Figure~\ref{fig:exifmapping}, the same blue color in the clothes and sky is decorated differently, with clothes more transparent and the sky deeper. This one-to-many mapping of the same color cannot be handled well if neglecting important shooting conditions. In the process of retouching, artists pay attention to the environment in which the photo was taken. In this research, those related information, such as exposure time and aperture, are introduced as a condition vector to get the shooting environment to handle the one-to-many mapping of colors.
\begin{figure}
\centering 
\begin{subfigure}[c]{0.27\textwidth}
\centering
\includegraphics[height=4cm, width=\textwidth]{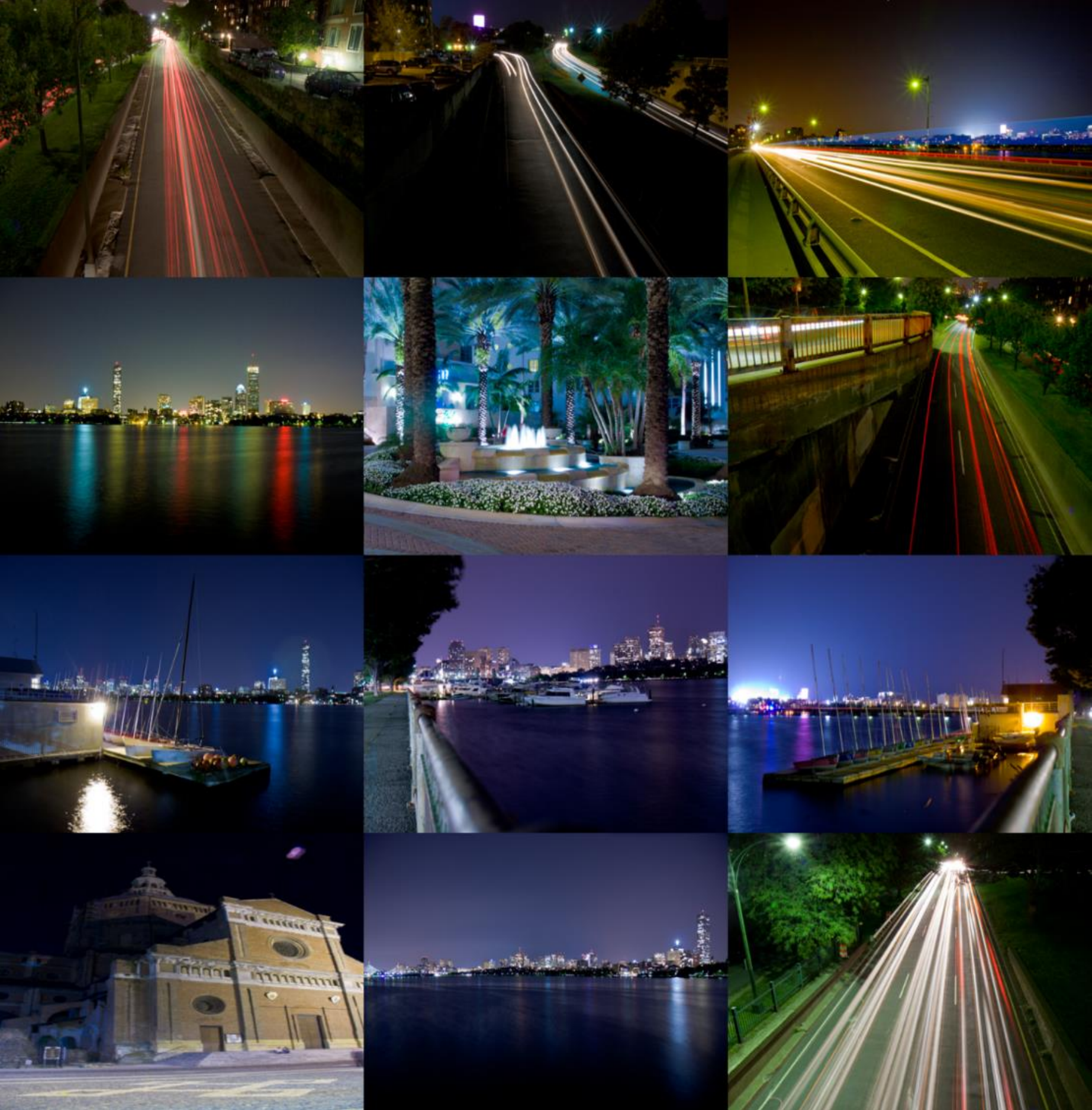}
\caption{Night Scene}
\label{fig:nightscene}
\end{subfigure}
\hfill
\begin{subfigure}[c]{0.17\textwidth}
\includegraphics[height=4cm,width=\textwidth]{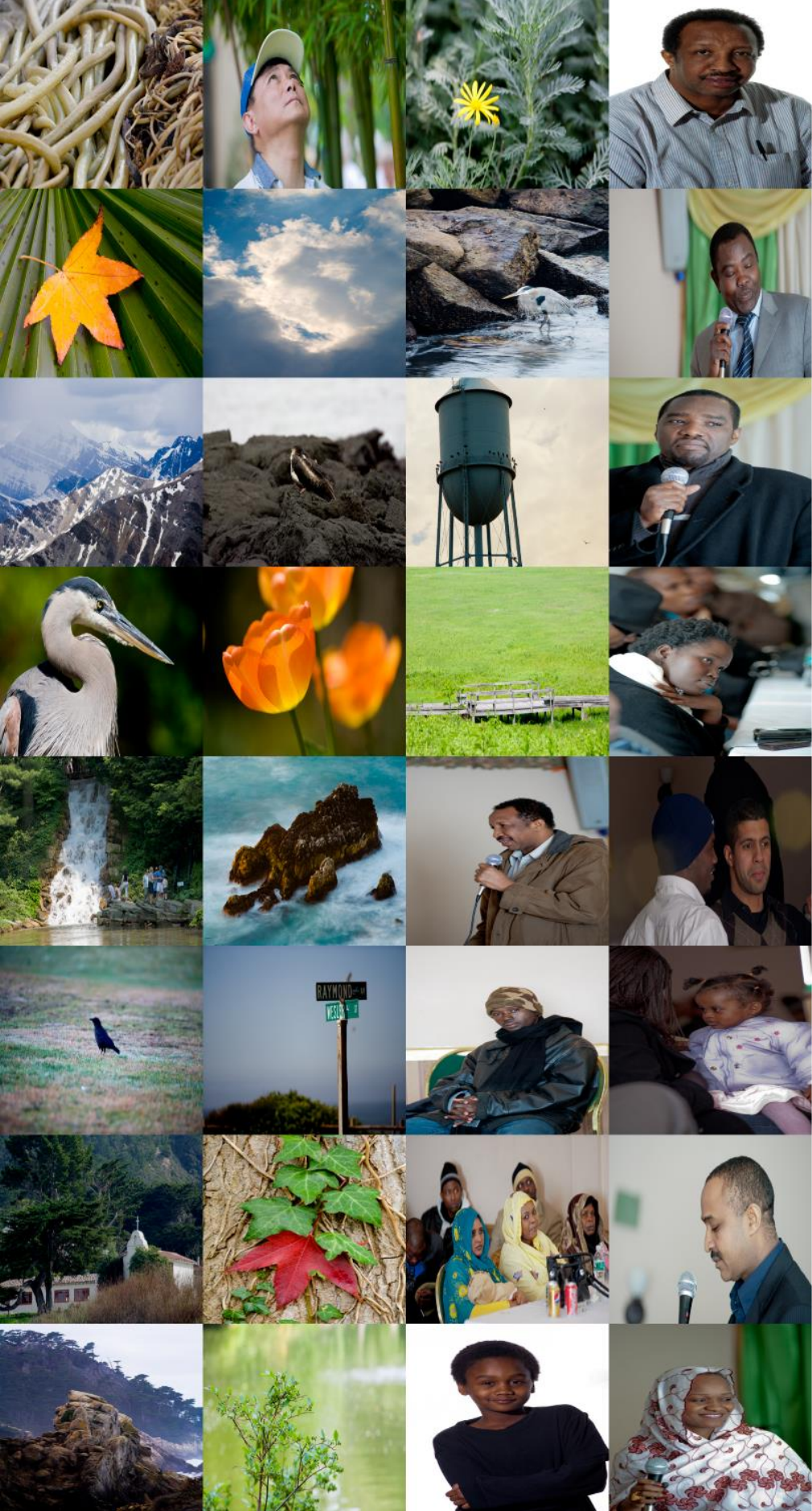}
\caption{Portrait}
\label{fig:potrait}
\end{subfigure}
\begin{subfigure}[c]{0.27\textwidth}
\includegraphics[height=4cm,width=\textwidth]{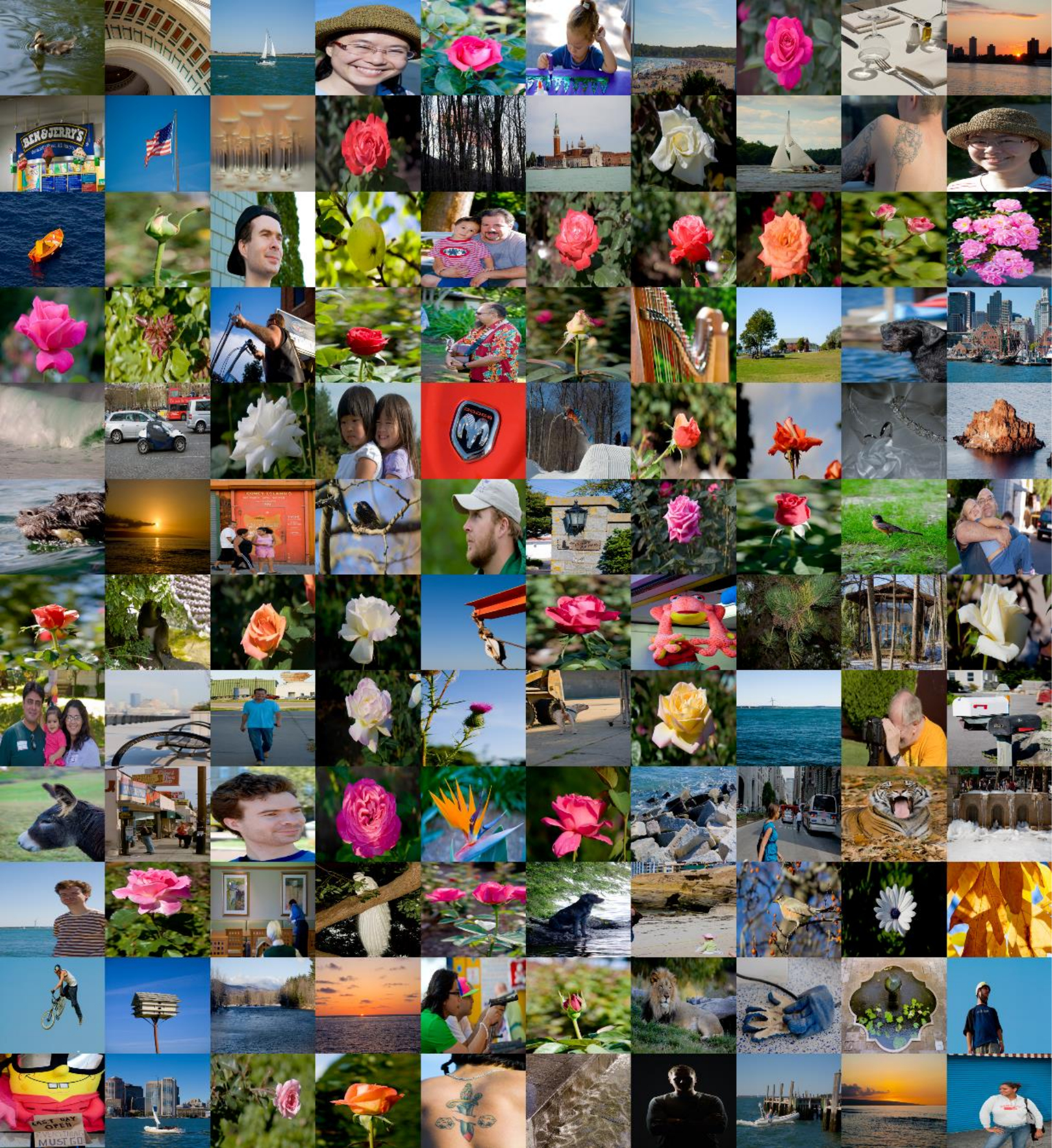}
\caption{Flowers}
\label{fig:flowers}
\end{subfigure}
\begin{subfigure}[c]{0.27\textwidth}
\includegraphics[height=4cm,width=\textwidth]{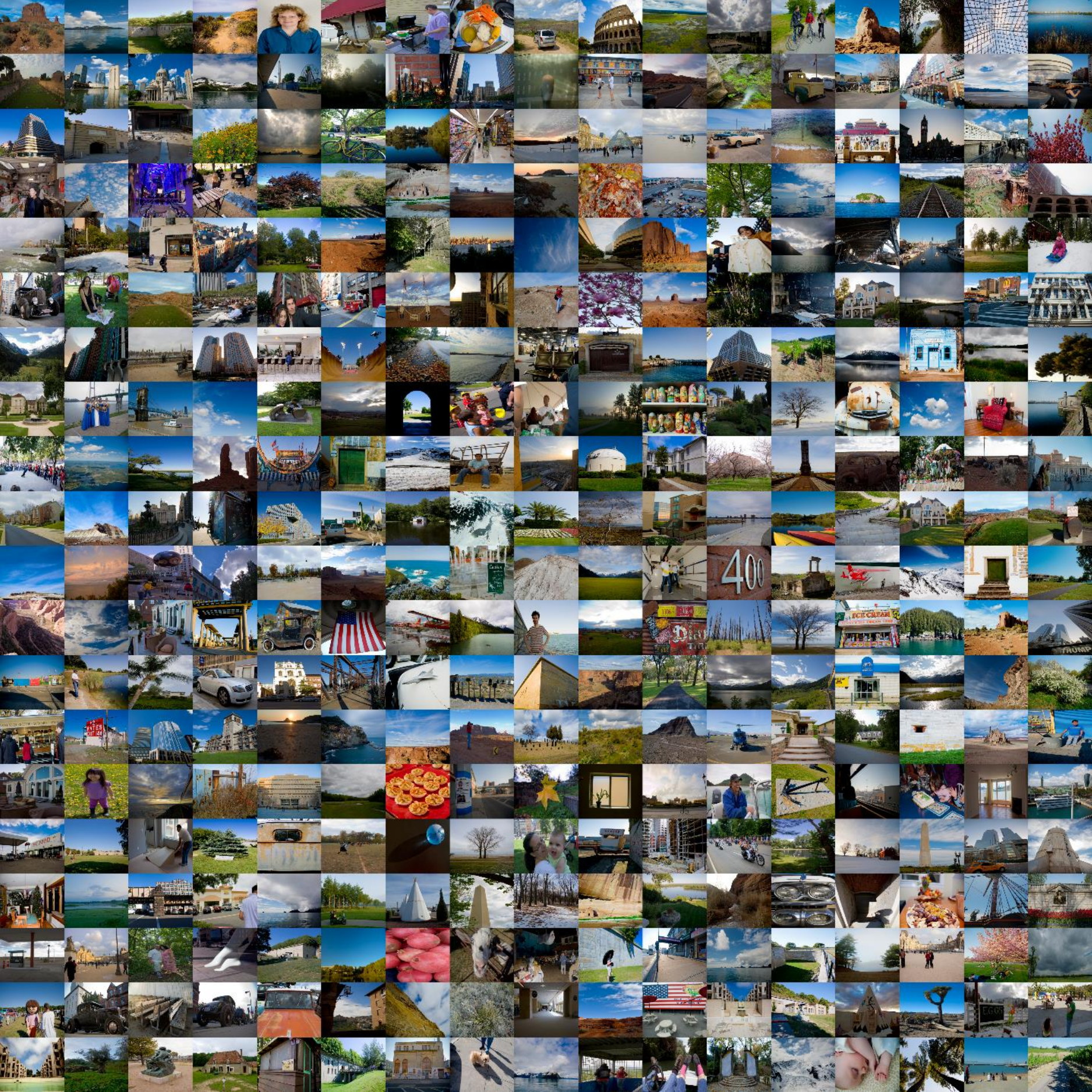}
\caption{Blue Sky}
\label{fig:bluesky}
\end{subfigure}
\caption{Images of some of our clustered categories}
\vspace{-0.3cm} 
\label{fig:cluster}
\end{figure}
Images to be processed are usually low-light and unsaturated in the chrominance plane. Gao et.al~\cite{gao2021real} verified that the retouching task can be decomposed into two independent parametric models to adjust brightness and color separately. TSN-CA~\cite{wei2021tsnca} uses a cascaded network to perform brightness enhancement and restoration, and the method focuses on low-light image enhancement. By researching the artist's retouching behavior, LCCNet uses a cascaded network to adjust the image to an appropriate brightness first, then the secondary network enriches color in the chrominance plane.

$\Delta E = \sqrt{(L_{1}-L_{2})^2+(a_{1}-a_{2})^2+(b_{1}-b_{2})^2}$ calculates the perceptual distance of the CIE Lab color space to measure the color accuracy. A Euclidean objective function is used~\cite{cheng2016Deep,dahl2016automatic} naturally based on this. However, the optimal solution to the Euclidean loss tends to the mean of set under a series of distinct $ab$ values~\cite{zhang2016Colorful}. In~\cite{zhang2016Colorful}, Zhang et.al researched 1.3M training images in ImageNet~\cite{russakovaky2015ImageNet} and found that natural images were usually with low $ab$ values and lack of high values. So using only L2 loss, the bias distribution of $ab$ values will result in mean desaturated images and cannot improve the class imbalance problem in the chrominance plane. We design the palette loss to improve the class imbalance problem by extracting color components in the hue plane and then fitting each color component independent of the number of pixels.

This paper has three main contributions:
\begin{enumerate}
    \item We introduce six pieces of information related to the shooting condition in the EXIF as a condition vector to help handle the one-to-many mapping of colors.
    \item We propose a two-stage network that brightens images first and then enriches them in the chrominance plane to simulate the artist's operations and makes our algorithm more similar to expert retouching.
    \item Hue palette loss function which balances color class in chrominance plane is proposed and makes images more saturated. 
  \end{enumerate} 
The remainder of the paper is structured as follows. The analysis of the dataset is described in Section~\ref{analysize}. Section~\ref{method} illustrates the methodology and details of LCCNet. Section~\ref{experiments} reports experimental results and comparative studies between competing methods as well as ablation studies. Section~\ref{conclusion} concludes the entire research.


\section{Analysis of the dataset}\label{analysize}

\subsection{Color correlation of EXIF}

One of the difficulties with image retouching is that one-to-many color mapping, and identically colored input may not correspond to the same output. As shown in Figure~\ref{fig:exifmapping}, the output blue of the sky is darker with the same blue input, because the bluer sky is preferred. Obtaining conditions in which the photograph is
taken (e.g.indoors and outdoors, portraits or landscapes) can help ease the one-to-many color dilemma in retouching, since darker blues are considered when the scene is known to be the sky. However, the shooting process is naturally a dimensional reduction process. While using a single image input, the shooting information has been discarded. Fortunately, camera devices record part of scene information, that is EXIF. EXIF of an image, including the imaging conditions, such as the aperture and exposure time. These conditions related to imaging can be used as condition input. For example, the EXIF of the portrait is different from that of sky, so the correlation between the EXIF and images' contents can be exploited to improve the one-to-many color mapping problem.

EXIF is used to record the attribute information and shooting data of digital photos, which can be attached to JPEG, TIFF, RIFF files, etc. After a photo is taken with an SLR or mobile phone, the camera generates an electronic file, such as JPEG, and saves it. Exposure Time, FNumber, Shutter Speed, Flash, Focal Length, Aperture, ISO Speed Ratings et.al, all these recorded imaging conditions are the EXIF of the corresponding picture. When taking photos, people will set shooting parameters based on the shooting scene, figure~\ref{fig:exifmapping} shows some photography habits: larger apertures and more portraits. Other than that, more exposure time means more light is admitted (usually used for darker scenes), et.al.

We performed an EXIF-scene correlation analysis on all 5000 images in the MIT-Adobe FiveK dataset. The Exposure Time, ISO, FNumber, Shutter Speed, Focal Length, and Bias Value related to scene information in EXIF are used to generate a six-dimensional vector. Clustering is widely used in the field of unsupervised data analysis, so it is used to analyze the aforementioned correlations. The MIT-Adobe FiveK dataset contains a wide variety of contents, with 5000 pictures including portraits, night scenes, blue sky, and other scenes. According to the vector, they are clustered into 30 categories. The result shows that there are 16 clusters with distinct styles. We select some of them to display in Figure\ref{fig:cluster}. As shown in Figure\ref{fig:cluster}(a), all images are night scenes, while in Figure\ref{fig:cluster}(b), the proportion of portraits is the highest. There are many red flowers in Figure\ref{fig:cluster}(c), while almost all pictures in Figure\ref{fig:cluster}(d) are blue sky scenes. From the foregoing, it appears that EXIF and scene are related. This correlation can help determine the scene style of the image. Naturally, EXIF can be added to the network to assist in image retouching.

\subsection{Retouching operations of artists}

\begin{figure}\scalebox{0.5}
\centering
\includegraphics[width=\textwidth]{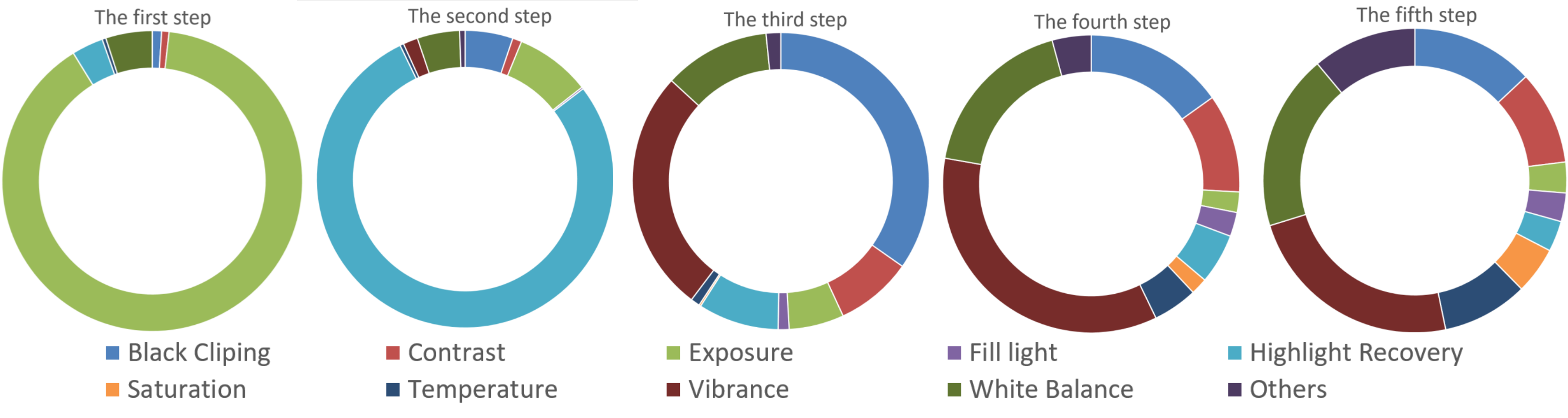}
\caption{Retouching operations of artists in the dataset}
\vspace{-0.3cm} 
\label{fig:steps}
\end{figure}

Artists' retouching behaviors should be taken into consideration since this machine learning problem is designed to mimic artists' photo retouching. The MIT-Adobe fiveK dataset records artists' all intermediate operations for retouching. As seen in Figure~\ref{fig:steps}, we count the artists' first five retouching steps of the former 500 images, and the first two most popular actions are Exposure adjustment and Highlight Recovery, which are used to adjust the brightness of the image. In the third step, Black Clipping with the highest proportion is used to restore the dark details after the brightness adjustment. Once the luminance is adjusted properly, the subsequent Vibrance and White Balance adjust the image in the chrominance plane to make the image more saturated and vibrant. 

Artists tend to adjust the brightness of the image first and then work on the chrominance plane. The main reason is that the compressed gamut range in low illumination and low luminance makes color adjustments difficult. As shown in Figure~\ref{fig:allab}(a), scatters are six main colors extracted from 5000 raw images, with a low-illumination condition. Chrominance distribution of the picture is aggregated, and it is difficult to distinguish and classify.

\subsection{Distribution of $ab$ values}

As mentioned above, natural images in ImageNet are usually with low $ab$ values and desaturated. Similarly, we perform $ab$ values distribution analysis for all 5000 images of Expert C in MIT-Adobe FiveK, which are most commonly used as GT for retouching tasks. For each image, the first six major colors in the CIE Lab gamut (we use CIE Lab gamut for analysis because it is designed to approximate human visual perception) are extracted and analyzed. In this gamut, the values of $ab$ closer to 0 indicates the lower in saturation, while close to 1/ -1 indicate the higher saturation. L value indicates the luminance of the image, and the value of -1/ 1 represents the darkest/brightest the image. 

\begin{figure}
\centering
\includegraphics[width=\textwidth]{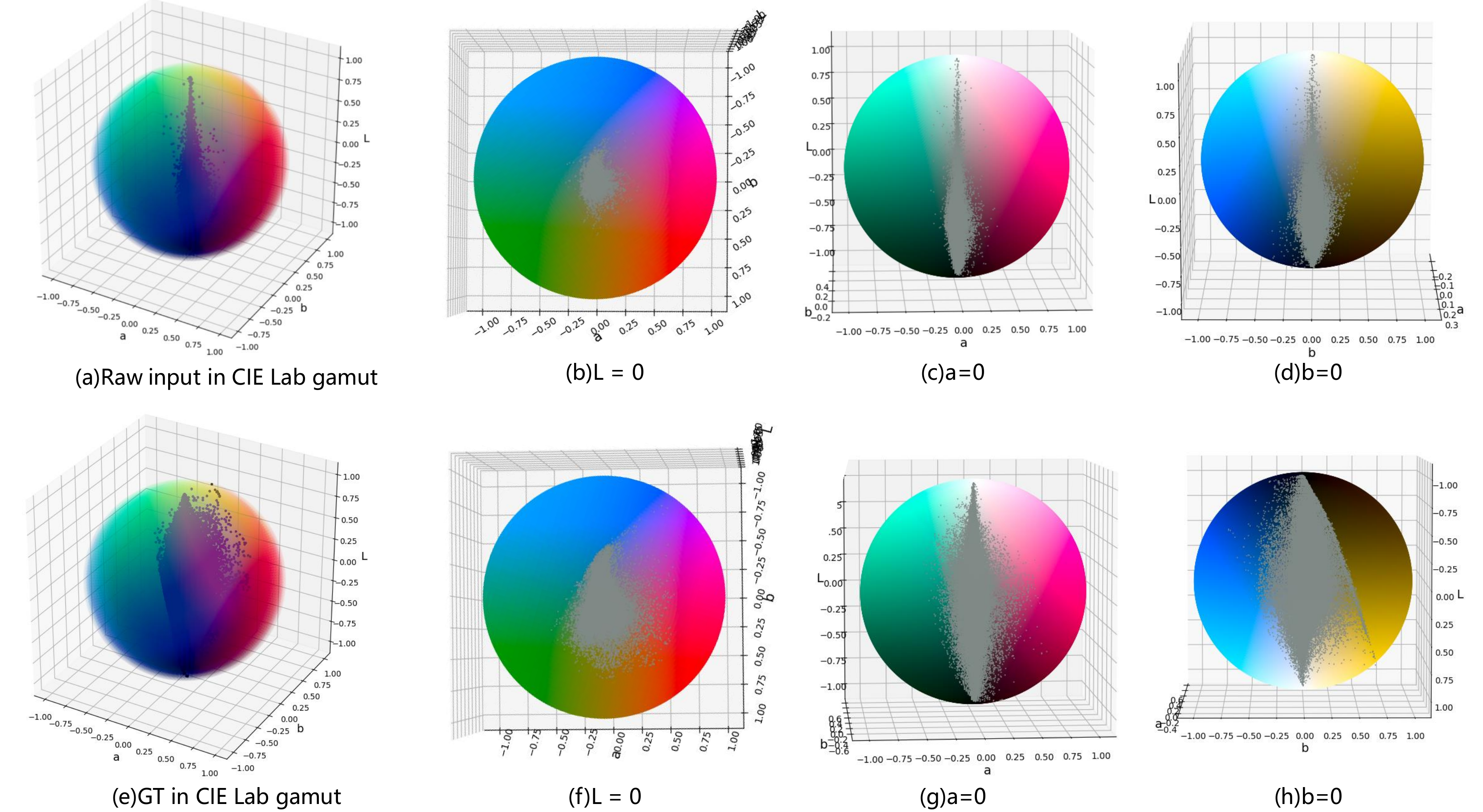}
\caption{Distribution of raw images and GT images color values in CIE Lab color gamut}
\vspace{-0.6cm} 
\label{fig:allab}
\end{figure}

As shown in Figure~\ref{fig:allab}(e), scatters are main colors of GT, the colored spheres represent the range of the CIE lab gamut. The closer to the surface of the sphere, the higher the saturation is, while the distribution of the dataset is concentrated inside the sphere. Using the L=0, a=0, and b=0 planes to cut the color gamut, $ab$ values of  GT concentrated in the area around a=0 and b=0, indicates that images have low saturation. 

\section{Method}\label{method}

In artful image retouching(AIR), common operations such as adjusting the exposure of highlights/ right hues/ dark hues/ shadows, or adjusting color temperature/hues by manipulating a global tone mapping curve only modify global features. Local operations using pixel-to-pixel color transformation are meticulous, which may cause degradation in spatial structures or inconsistent tone style in the local area. As mentioned above, we counted the first 500 images of artists' retouching operations in the dataset. All 500 images used only global operations such as mapping curves. That is why LCCNet solely focuses on global operations and yields impressive results.

AIR is a cognitive task associated with personal aesthetics, not merely a signal processing problem like image super-resolution. The aesthetic evaluation of the image itself is vague, but artists can give a reason for their every operation in the process of retouching, such as "Hi, I think the background is too dark, I have to brighten it up". Therefore, not only should we focus on the retouched image, but also the retouching behavior should be considered. First, EXIF of photos, which has been ignored in previous methods is introduced. Secondly, the operation of artists' manual retouching is concerned and imitated. Finally, rebalancing the color imbalance in the dataset. Based on the above analysis, the EXIF is used as condition input to improve the one-to-many color mapping problem. A two-stage network that cascade luminance and chrominance is designed. The hue palette loss is designed to solve the problem of uneven color distribution in the training dataset so that colors in an image get the same attention.

\subsection{Prior knowledge-EXIF}

It is natural to use EXIF for AIR to improve the automatic retouching algorithm since AIR is closely related to the image's scene. 

In~\cite{he2020csrnet,he2020modulation}, they use condition vectors generated from image content to adjust feature maps globally. Similarly, we use the EXIF condition vector to generate adjustment coefficients and apply them to feature maps. Because the scene information is associated with the whole image, the adjustment coefficients are global, not local.   

\begin{figure}
\centering
\includegraphics[width=\textwidth]{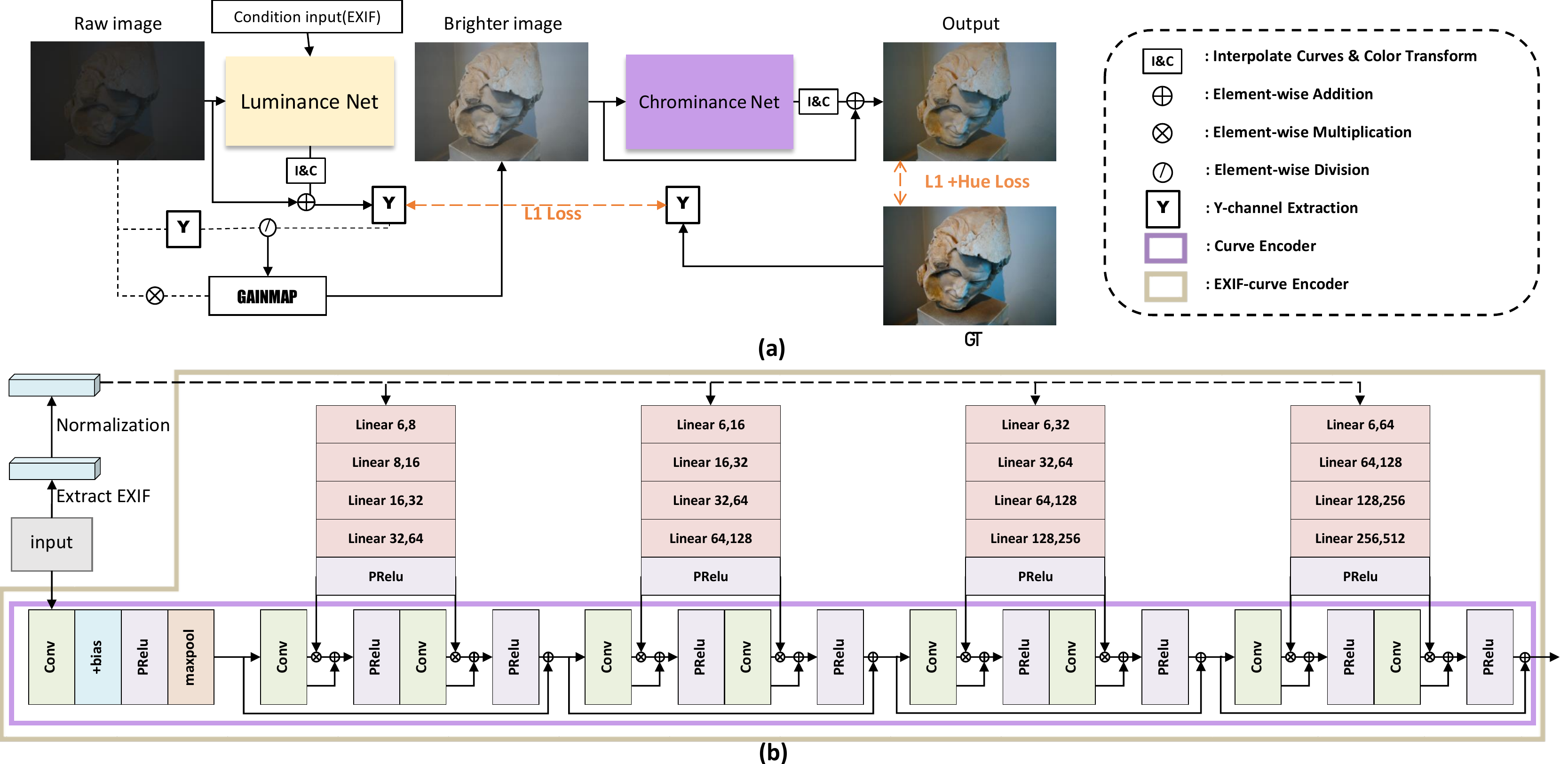}
\caption{Main structure of LCCNet}
\vspace{-0.5cm} 
\label{fig:network}
\end{figure}

As shown in the Figure~\ref{fig:network}(b), the condition vector passes through a series of fully-connected layers to generate global adjustment coefficients, which are applied to the feature map of each layer. 

\subsection{Luminance-chrominance cascading structure}


A cascaded two-stage network structure is obtained by simulating artists' retouching operations based on the above analysis, which adjusts brightness first then chrominance. As shown in Figure~\ref{fig:network}(a), the first-stage network focuses on illumination improvement, and the second-stage network is used to enrich the color of the highlighted image generated by the first-stage network. 

The luminance adjustment of the former stage network is implemented in the $YUV$ (BT601) color gamut. In this color gamut, the $Y$ channel indicates the luminance of the picture. The conversion between the $RGB$ color gamut and $YUV$ color gamut is linear. For input pixel $I(x, y)(I\in[0, 1], x, y\in H\times W  )$ in an image $I$ with $H \times W$ resolution, the conversion formula is as follows:
\begin{equation}
\left[\begin{array}{ccc}
    I(x,y)^{Y} \\
    I(x,y)^{U} \\
    I(x,y)^{V} \\
    \end{array}
    \right]=
\left[\begin{array}{ccc}
    0.299&0.587 &0.114\\
    -0.169&- 0.331 &0.5\\
    0.5&- 0.419 &- 0.081\\
    \end{array}
    \right]
\left[\begin{array}{ccc}
    I(x,y)^{R}\\
    I(x,y)^{G}\\
    I(x,y)^{B}\\
    \end{array}
    \right].
\end{equation}
$Y'$ indicates the luminance for whole image after brightness changes, and $gainmap$ can be expressed as $Y'$ element-wise division by $Y$. Here the $gainmap$ is a matrix with the same height and width as the input image. Each element in the matrix independently adjusts the pixel value of the corresponding position. Using $gainmap$ element-wise multiplies $RGB$ input to obtain the $R'G'B'$ after luminance adjustment.

Compared with the independent learning process in which all three $RGB$ channels are involved, using the same gain for each channel does not damage the proportion relationship of the three $RGB$ channels of the image itself, that is, the hue of the image. For the training process, this achieves luminance alignment without changing hue during adjustment.

The output of the first-stage network is used as the input of the second-stage network. Compared with the original input of low brightness, the brighter image input to the secondary network has a wider range of color gamut. In this case, the operation of chrominance becomes easier.

\subsection{Hue palette loss}

Using the Euclidean or L1 loss in photo retouching tends to average the set under a series of distinct $ab$ values, thus the colors occupying larger areas of the graph will be paid more attention and the smaller parts may be ignored. This is contrary to the usual cognition because it is not the part with the largest area in the image which attracts the most attention, but the part with a significantly different color from the surroundings that attracts people's attention, no matter how small its area is. According to our analysis, the dataset presents a low ab value and low saturation distribution, thus using L1/L2 loss alone may ignore saturated portions with high ab values and inevitably lead to a less vibrant output.

\begin{figure}
\centering
\vspace{-0.3cm} 
\includegraphics[width=\textwidth]{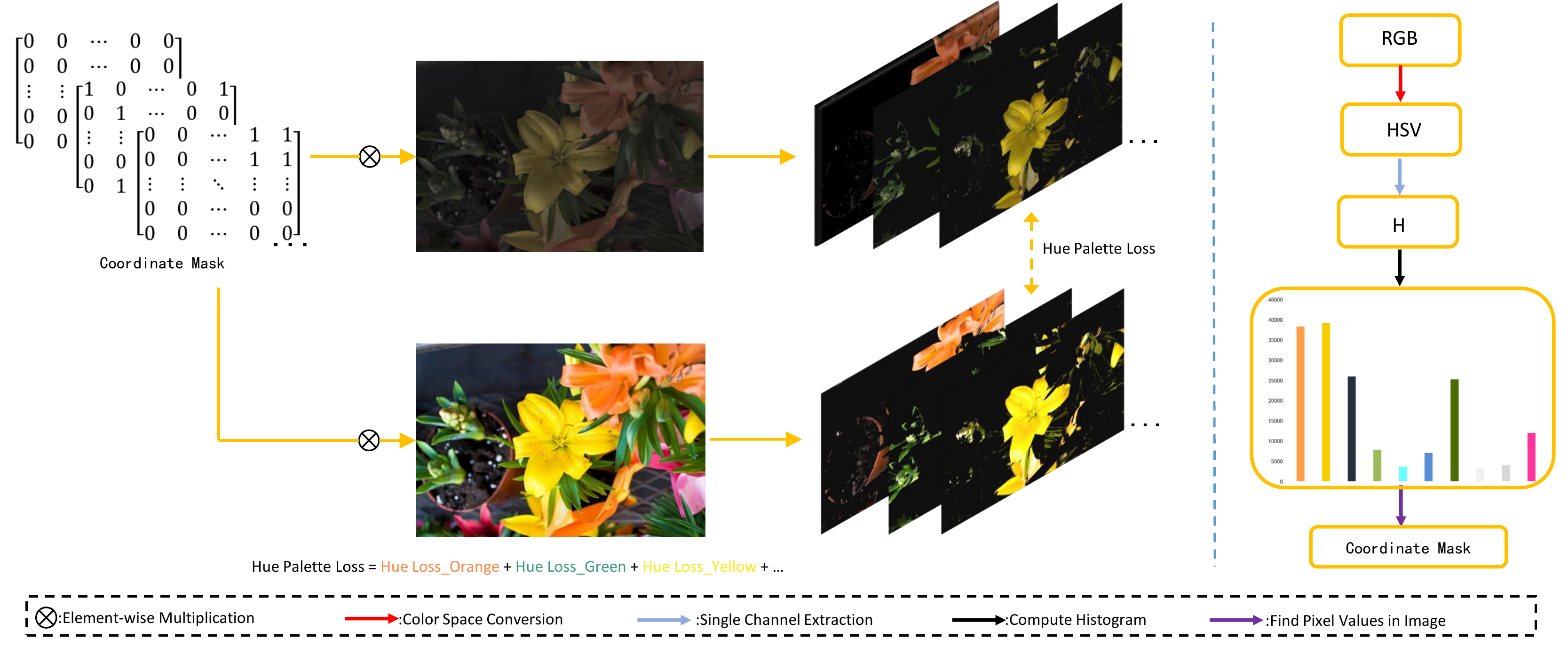}
\caption{The generation mode of Hue palette loss. All graphs are real calculation processes, and the color of the histogram is the corresponding color in the original image.}
\vspace{-0.5cm} 
\label{fig:hueloss}
\end{figure}

Hue palette loss is designed to balance the attention of each color component to saturate the image more. As shown in Figure~\ref{fig:hueloss}, first, the $RGB$ input is converted to the $HSV$ gamut, and then the $H$ channel is extracted because the $H$ channel is hue and represents the category of color contained in the image. The histogram of $H$ channel is exploited. In this histogram, too wide of bin width can cause color confusion, while too narrow will cause similar colors to be divided into different ranges. By measuring different interval widths and computational costs by experiment, we divide the H channels in the range of [0, 360] into 10 equal bins, each interval width is 36. More details of can be seen in supplementary material.

Each bin represents a color in the image and the position of each color in the image can be obtained by reversely searching for the coordinates of the corresponding pixel value in the image. The coordinate points in the image corresponding to the current color are assigned to 1; otherwise, the coordinates are assigned to 0. The Coordinate Mask is used to element-wise multiply input and the GT image. Hue loss for the corresponding color is obtained and normalized by using the number of pixels for the corresponding color to eliminate the impact of the area. The Hue palette loss is formulated below:
\begin{equation}\label{eq:totalhueloss}
\begin{aligned}
Hue\_palette\_loss &= \sum_{j=1}^k Hue\_loss_{k}&= \sum_{j=1}^k \frac{d_{1}(output*CM_{j},GT*CM_{j})}{SUM(CM_{j})}.
\end{aligned}
\end{equation}
Here in Equation(\ref{eq:totalhueloss}) $CM_{j}$ indicates the j-th of Coordinate Mask, and $*$ means Hadamard product. For two $m \times n$ matrices $A=[a_{ij}]$ and $B=[b_{ij}]$,  $(A*B)_{ij} = a_{ij}b_{ij}$, $d_{1}(A,B) = \sum_{i=1}^n\sum_{j=1}^n|a_{ij}-b_{ij}|$,
$SUM(A) = \sum\sum a_{ij}$.

Hue palette loss eliminates the influence of the color area by normalizing the number of pixels corresponding to the color and colors of different areas get the same attention. In this way, imbalances in the dataset due to low $ab$ values are resolved and the images are more saturated.





\section{Experiments and implementation details}\label{experiments}
\subsection{Dataset and Metrics} 
MIT-Adobe FiveK dataset is a commonly-used photo retouching dataset that consists of 5,000 raw photographs taken with SLR cameras. All these photographs have been retouched by five retouching experts (A/B/C/D/E). Following the previous methods ~\cite{he2020csrnet,Song2021starenhancer,kim2020globallocal,moran2021CURL,gharbi2017hdrnet,yang2022AdaInt,wang2019DUPE,zeng20203Dlut,moran2020deeplpf} to use the retouched results of expert C(sRGB color gamut) as the ground truth (GT), and adopting the same pre-processing procedures as ~\cite{he2020csrnet,hu2018whitebox,yu2018deepexposure} that all the images are resized to 500px on the long edge for training. The images of the default input style as input are used, and the first 4500 image pairs are used for training and the last 500 image pairs for testing. PSNR, SSIM, and LPIPS~\cite{zhang2018lpips} are used to evaluate the performance of all methods.

\subsection{LCCNet}


The curve approach in~\cite{Song2021starenhancer}, named basic enhancer is applied as the basic model in LCCNet. The basic enhancer assumes that each channel of the output image is correlated with each channel of the input image as well as the coordinate maps ${x, y}$ of each image. We used the simple form of the basic enhancer, where the output channels related to the coordinate maps ${x, y}$ are affected by the image size but independent of the input pixel values. We implemented LCCNet using the pytorch framework~\cite{paszke2019pytorch}. LCCNet is a two-stage network, images with EXIF(Very few images contain some but not all of the six pieces of information. For information not included in the EXIF, we chose to set it to 0.) are input to the network together. The $Y'$ channel generated after the Brighter Net training uses the L1 loss to fit the $Y$ channel of the GT image. $Gainmap$ is applied to the original input by multiplying the element direction to generate a highlight image for Richer Color Net. Hue palette loss and L1 loss under the CIE-Lab gamut are used for the final output with the same weight. 

\subsection{Comparison with State-of-the-Arts}

LCCNet is trained on a single NVIDIA V100 for 2000 epochs with a batch of 32 for 15h. In addition, the random seed is set to 6 to ensure that the result can be reproduced. LCCNet is compared with contemporaneous methods on MIT-Adobe FiveK as shown in Table\ref{table:comparsion}. HDRNet, DPE, DUPE, DeepLPF are local based method. CURL, STAREnhancer and our method are globaled based and use curve-based approachs. A3D-LUT and AdaInt are both tabel-based and CSRNet use condition vector generated from images for global retouching. LCCNet achieves optimal PSNR and SSIM in both low and original resolution. At the same time, the best SSIM is in low resolution and the second in high resolution. Moreover, because our method is a global adjustment, the effect is stable with resolution varying. 

\setlength{\tabcolsep}{4pt}
\begin{table}
\begin{center}
\caption{Quantitative comparison with contemporaneous methods on MIT-Adobe-FiveK. {\color{red}Red} figure indicates the best and {\color{blue}Blue} indicates the second best. Note that for \underline{underlined} methods we retrained them, and for other methods we use the result in their origin paper (some are absent (“/”) for lacking of source code).}
\label{table:comparsion}
\begin{tabular}{ccccccc}
\hline\noalign{\smallskip}
 & &480P & & & Full resolution & \\
Method & PSNR(↑) & SSIM(↑) &LPIPS(↓)& PSNR(↑) & SSIM(↑) &LPIPS(↓)\\
\noalign{\smallskip}
\hline
\noalign{\smallskip}
\underline{HDRNet}~\cite{gharbi2017hdrnet} &24.66dB &0.915 & 0.076&23.20dB& 0.917&0.120  \\
DPE~\cite{chen2018dpe} &23.75dB  &0.908 &/ &22.15dB&0.850 & /\\
DeepUPE~\cite{wang2019DUPE} &/&/&/& 23.24dB  & 0.893 & 0.158\\
DeepLPF~\cite{moran2020deeplpf} &23.93dB &0.903   &0.582&24.48dB &0.887 &0.103\\
\underline{CSRNet}\cite{he2020csrnet} &25.14dB  & 0.924 &0.062 & 24.82dB & 0.926 &0.091 \\
A3D-LUT~\cite{zeng20203Dlut} &{\color{blue}25.50dB}& 0.890 & 0.051 &23.17dB&0.864&0.145\\
CURL~\cite{moran2021CURL} &/ &/ &/ &24.20dB& 0.880& 0.108\\
\underline{STAREnhancer}\cite{Song2021starenhancer} &25.40dB& {\color{blue}0.939}  & {\color{blue}0.044} &25.46dB&{\color{red}0.948}&{\color{blue}0.083}\\
AdaInt~\cite{yang2022AdaInt} &25.49dB&0.926  &/ &{\color{blue}25.48dB}&0.934 &/ \\
\hline
LCCNet &{\color{red}25.81dB} &{\color{red}0.944}  &{\color{red}0.041} &{\color{red}25.80dB}  &{\color{blue}0.947}  & {\color{red}0.074}\\
\hline
\end{tabular}
\end{center}
\end{table}
\setlength{\tabcolsep}{1.4pt}

\begin{figure}
\centering
\vspace{-0.3cm} 
\includegraphics[width=\textwidth]{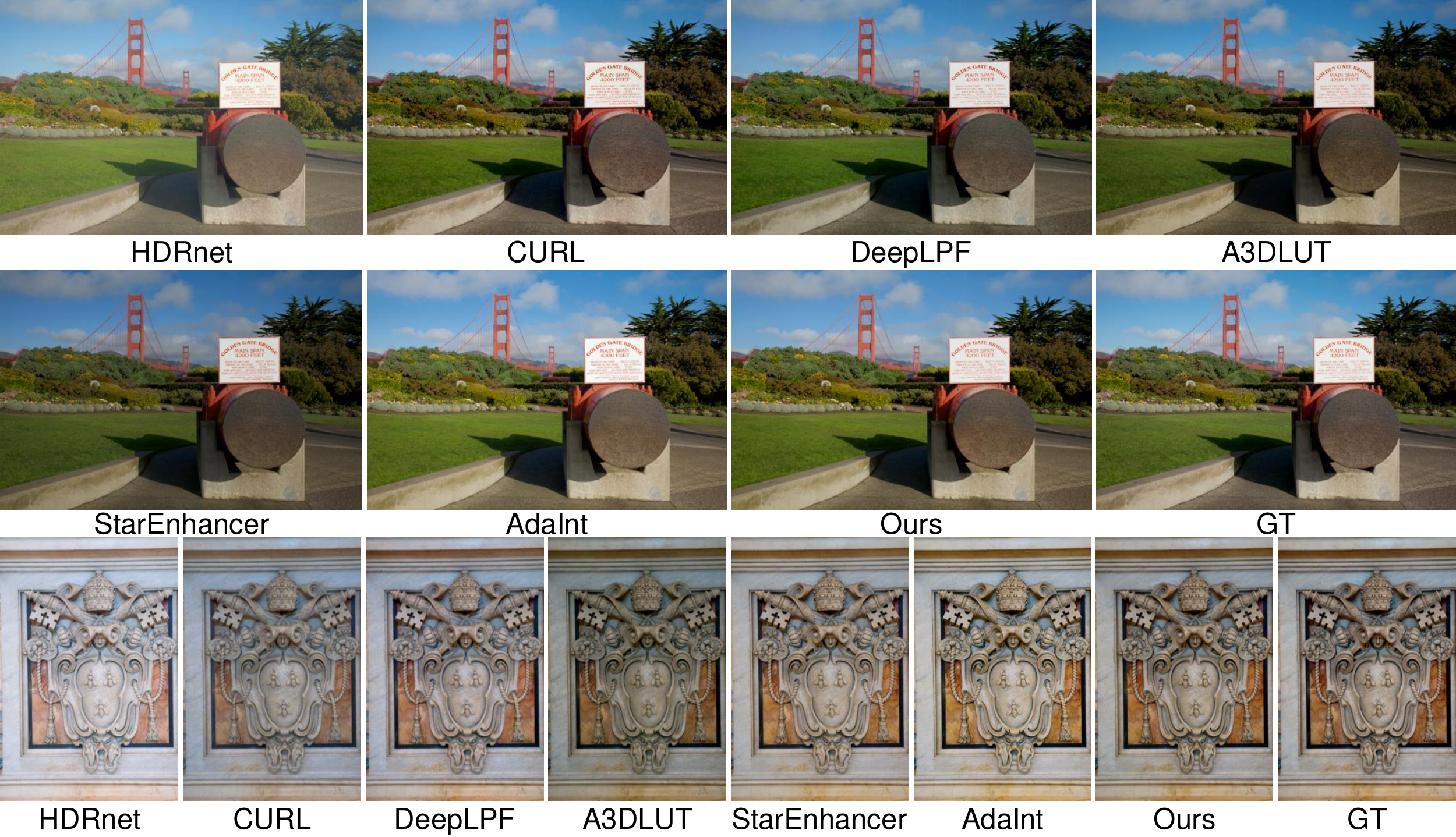}
\caption{Qualitative comparison with contemporaneous methods on MIT-Adobe fiveK.}
\vspace{-0.3cm} 
\label{fig:comparsion}
\end{figure}

Figure\ref{fig:comparsion}  shows some of the test images. Thanks to the design of the luminance network, in the first scene LCCNet's luminance is closest to the real scene. In addition, LCCNet also performs well for the red color in a small area thanks to hue palette loss. The second scenario has a lot of interlaced details, and our method successfully focuses on each area compared to other methods. Please see the supplementary for more comparisons.


\subsection{Ablation study}

To demonstrate the effectiveness of different components of LCCNet, several ablation studies are conducted on the MIT-fiveK dataset. As shown in Table\ref{table:ablation}, each component in LCCNet can independently improve the performance of the basic model and work together. 

\begin{figure}[!h]
\centering
\includegraphics[width=\textwidth]{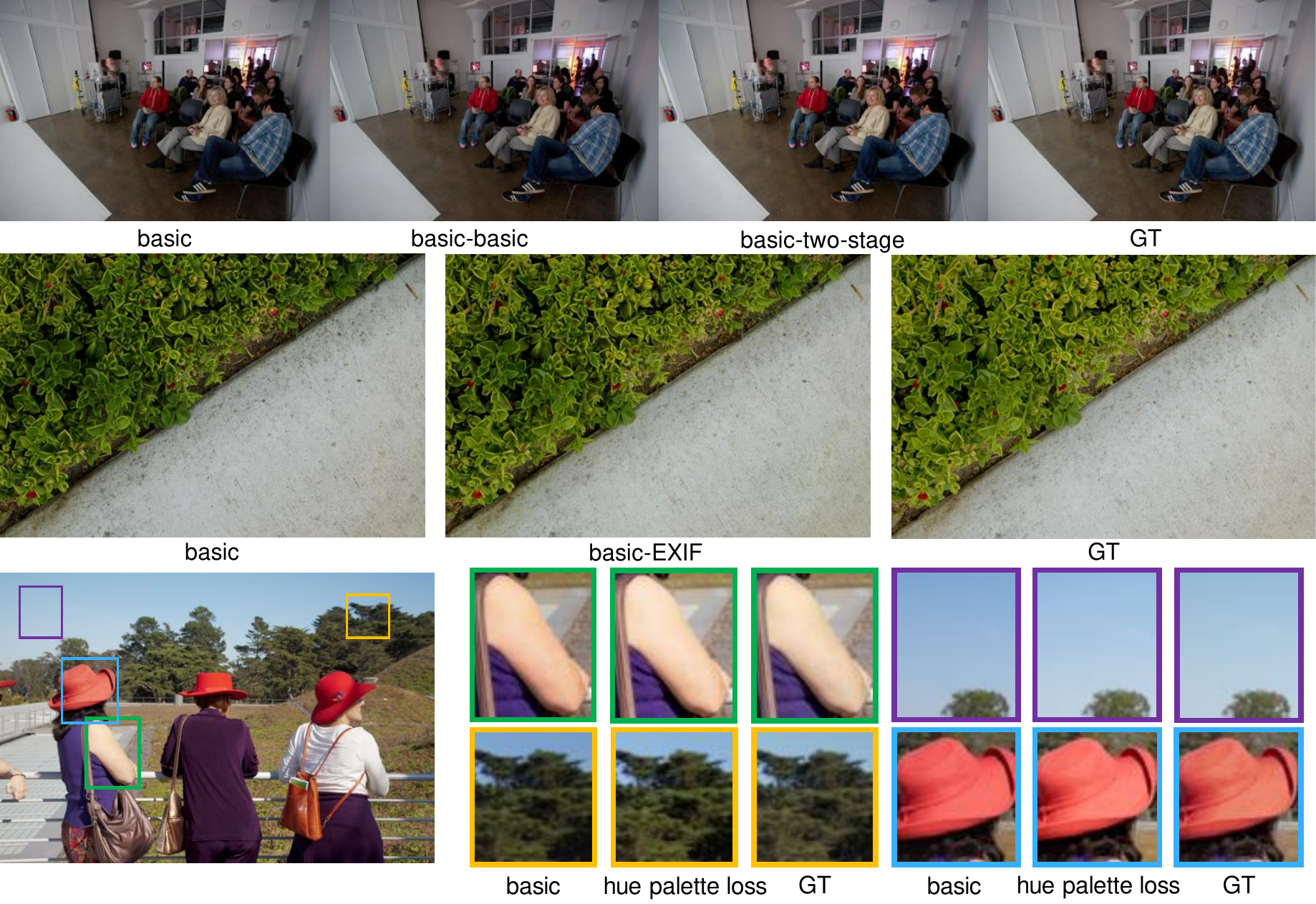}
\caption{The effect of basic model and adding different modules that we proposed.}
\vspace{-0.3cm} 
\label{fig:ablation}
\end{figure}

Noting that using our proposed $gainmap$ structure performs better than directly cascading two basic models, because the two networks perform their own independent functions. As shown in the top row of Figure \ref{fig:ablation} it's obvious that the clothes generated by basic model is darker. Using the  extracted from GT and raw images, then apply them to raw inputs. The brighter images gain from $gainmaps$ inputted to basic model can achieve a highly 29.79 dB of PSNR. It can be seen that after adjusting the brightness to the appropriate range, the subsequent color processing is easier and more effective. 
In the second row in Figure \ref{fig:ablation}, EXIF is used to map colors correctly and green is closer to the grass in nature. In addition, since CSRNet~\cite{he2020csrnet} is a typical method that uses condition input for image retouching, we apply EXIF as a condition input to CSRNet in the same way as LCCNet, and Table \ref{table:csrnet} shows a steady increase in PSNR and SSIM at the cost of very few parameters. Details of the experiments and comparisons of the effects are provided in the supplementary materials.

Hue palette loss is designed to focus on fewer colors in unbalanced datasets, in the last row of  Figure \ref{fig:ablation}, the white arm position is not affected by a dark background, and the blue saturation of the sky is higher and closer to reality.

\setlength{\tabcolsep}{4pt}
\begin{table}
\begin{center}
\caption{Different components of LCCNet(480P)} 
\label{table:ablation}
\begin{tabular}{ccccccc}
\hline\noalign{\smallskip}
Method &basic model&basic-EXIF&basic-two stage&basic-basic&basic-hue loss&LCCNet\\
\noalign{\smallskip}
\hline
\noalign{\smallskip}
PSNR(↑) &25.40dB& 25.73dB& 25.72dB&25.58dB& 25.69dB& 25.81dB \\
SSIM(↑) & 0.939& 0.939 & 0.942 &0.940 &0.941 & 0.943 \\
\hline
\end{tabular}
\end{center}
\vspace{-0.5cm}
\end{table}

\setlength{\tabcolsep}{4pt}
\begin{table}
\vspace{-0.3cm}
\begin{center}
\caption{original CSRNet and CSRNet-EXIF} 
\label{table:csrnet}
\begin{tabular}{ccccc}
\hline\noalign{\smallskip}
Method & PSNR(↑) & SSIM(↑) & Parameters(↓)\\
\noalign{\smallskip}
\hline
\noalign{\smallskip}
CSRNet  & 23.69dB & 0.895  & 36489 \\
CSRNet-EXIF & 23.94dB & 0.896  & 37406 \\
\hline
\end{tabular}
\end{center}
\vspace{-0.7cm}
\end{table}
\setlength{\tabcolsep}{1.4pt}

\setlength{\tabcolsep}{1.4pt}

\section{Conclusion}\label{conclusion}

In this paper, we propose LCCNet which focuses more on artists' operations and achieve state-of-art performance on the MIT-Adobe-FiveK. We add EXIF to an automated retouching network, which proves its effectiveness on LCCNet and CSRNet. At the same time, the luminance-chrominance decoupling of humanoid operations is effective. Finally, a balanced hue palette loss for solving the imbalance in the training dataset is proposed. Noting that our approach is generic and therefore can be extended to other methods. 

There is still room for further experimental optimizations of the optimal EXIF information combination. A two-stage network brings more computational cost compared with benefits, a better way of cascading need further research.

\section{Appendix}

\subsection{EXIF in CSRNet}\label{sec. exifincsr}

In LCCNet, EXIF has been proven to work in AIR, and naturally, it can also be added to other auto retouching networks to make the output more in line with human aesthetics. In this section, EXIF is added to CSRNet\cite{he2020csrnet}. Similarly, a noticeable improvement in the same condition as their paper set is got.

\begin{figure}[htbp]
\centering 
\begin{subfigure}[c]{0.49\textwidth}
\centering
\includegraphics[width=\textwidth]{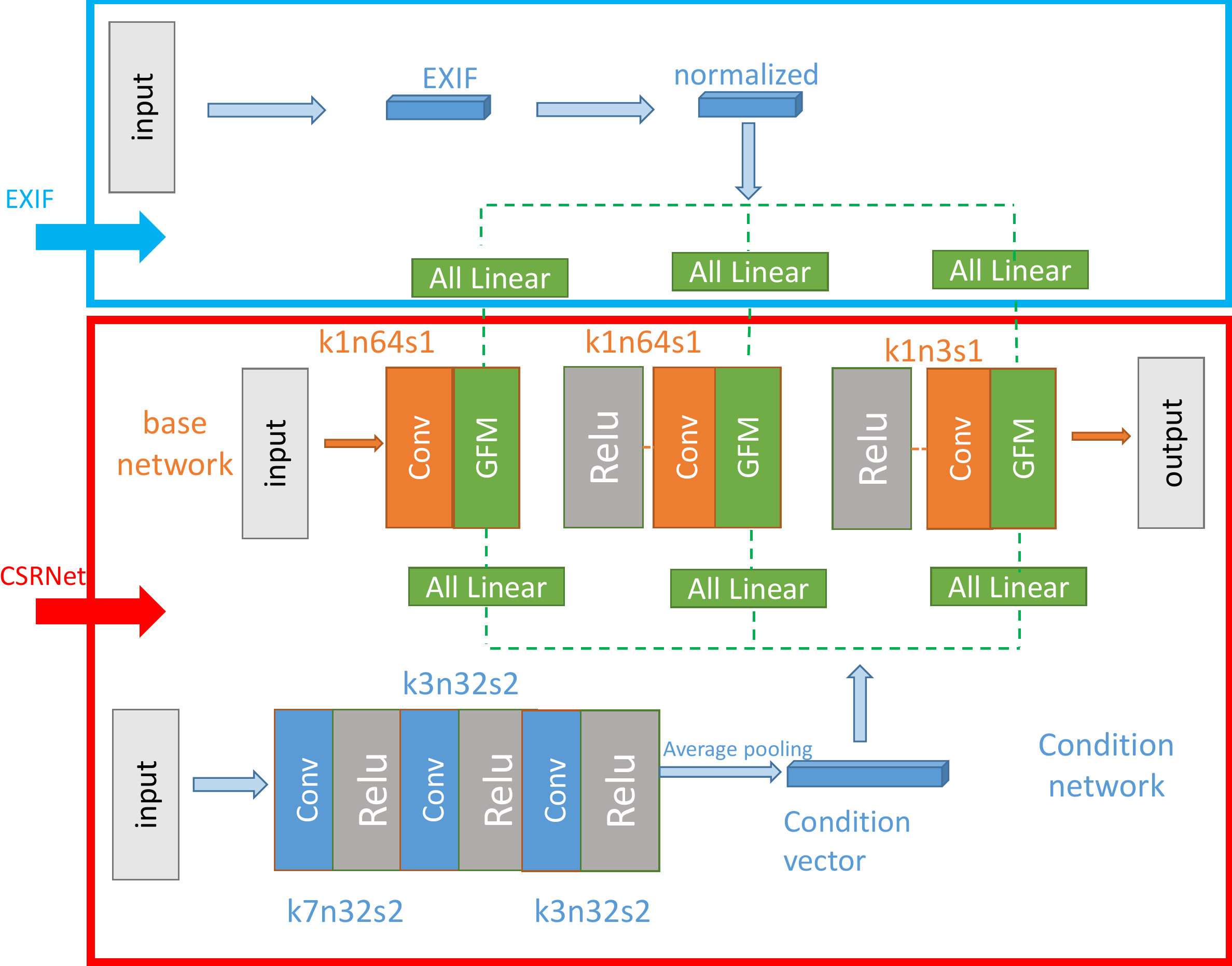}
\caption{Main Structure of CSRNet and Method of Adding EXIF}
\label{fig:exifincsr}
\end{subfigure}
\hfill
\begin{subfigure}[c]{0.49\textwidth}
\includegraphics[width=\textwidth]{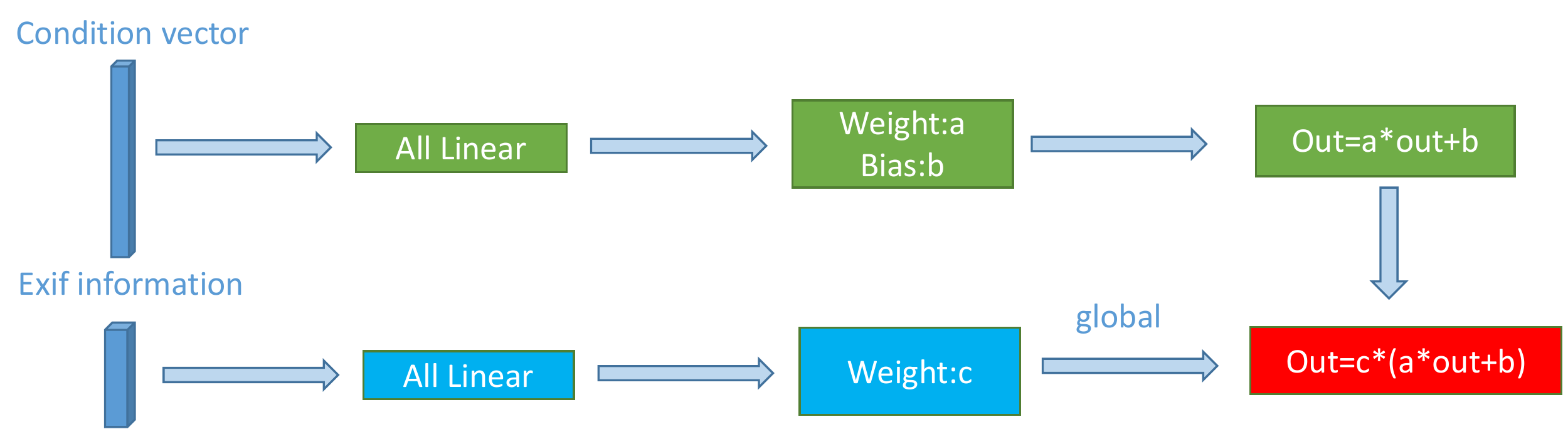}
\caption{EXIF Generating Global Coefficient $c$}
\label{fig:global-c}
\includegraphics[width=\textwidth]{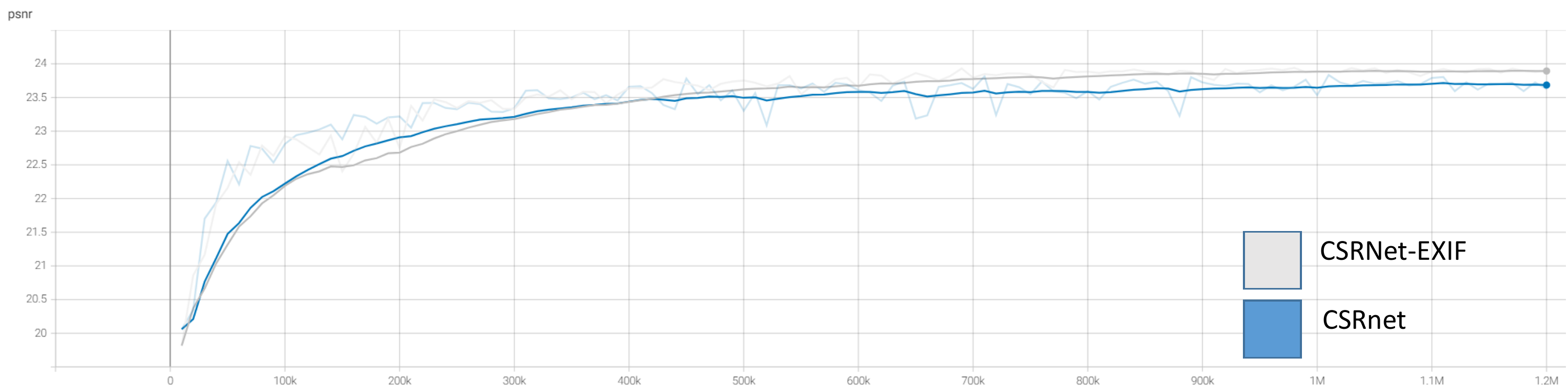}
\caption{EXIF Test Set PSNR Curves for CSRNet and CSRNet-EXIF (after smoothing).}
\label{fig:psnr-curve}
\end{subfigure}
\caption{Original CSRNet and CSRNet-EXIF}
\end{figure}

CSRNet consists of a base network and a condition network, and it uses global coefficients obtained from conditional networks to  modulate the intermediate features. As shown in the Figure \ref{fig:exifincsr}, since CSRNet regards image's content as a condition vector, the scene information contained in the EXIF can also be used as a condition vector. A global adjustment coefficient $c$  generated by EXIF is added to the original CSRNet affine transformation.

\setlength{\tabcolsep}{4pt}
\begin{table}[htbp]
\begin{center}
\caption{original CSRNet and CSRNet-EXIF} 
\label{table:csrnet-appendix}
\begin{tabular}{ccccc}
\hline\noalign{\smallskip}
Method & PSNR & SSIM & FPS & Parameters\\
\noalign{\smallskip}
\hline
\noalign{\smallskip}
CSRNet  & 23.69dB & 0.895 & 12.9 & 36489 \\
CSRNet-EXIF & 23.94dB & 0.896 & 13.5 & 37406 \\
\hline
\end{tabular}
\end{center}
\end{table}
\setlength{\tabcolsep}{1.4pt}

\begin{figure}
\centering
\includegraphics[width=\textwidth]{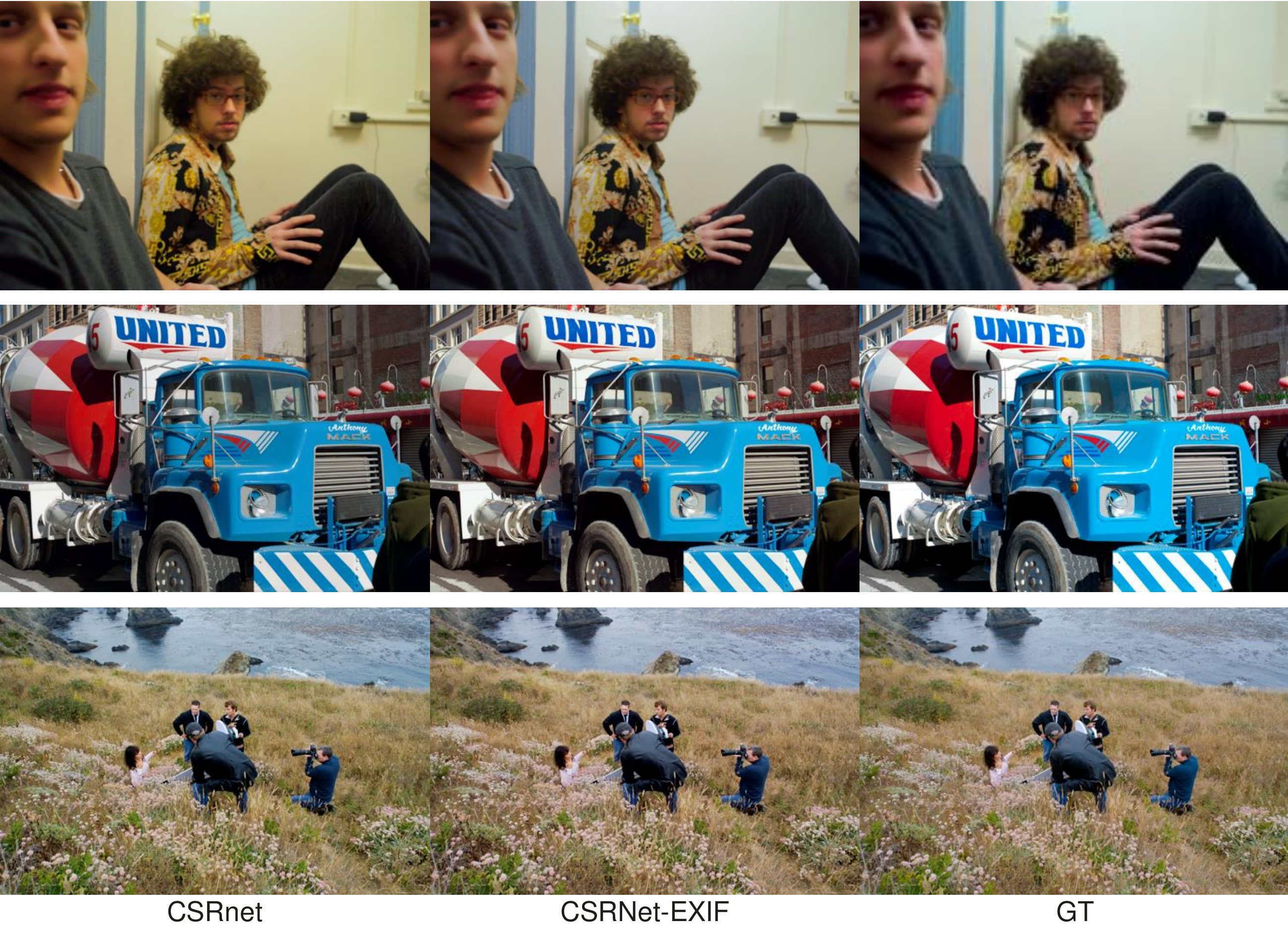}
\caption{Qualitative comparison between CSRNet and CSRNet-EXIF.}
\vspace{-0.6cm} 
\label{fig:csr-csrexif}
\end{figure}

Under the same training parameters and dataset partitioning method as CSRNet (the dataset paper released is used), the introduced prior knowledge increases 2.5\% of parameters, increases 0.24 dB PSNR stably, decreases 5.1\% MSE, and accelerates the inference speed of the model. Although the number of network parameters is slightly increased, the inference process is easier for the network due to the increase of effective information, which accelerates the inference process of the network. The curve in the Figure \ref{fig:psnr-curve} shows that the PSNR increases stably in the test set. And Figure\ref{fig:csr-csrexif} shows that the introduction of EXIF priori knowledge significantly improved the color effect of some scenes on CSRNet that are strongly related to the scene information.

The six-dimensional vector used is $[ISO, Exposure, Fnumber, Speed, Flength, Biasvalue]$, the average of the six-dimensional vector full dataset and the average of the training set are respectively $[456.131, 0.2423, 6.4261, 7.0484, 55.9858, 0.02]$ and $[441.23, 0.258, 6.4664, 7.0714, 56.6272, 0.018]$. In order to make the influence of each dimension the same, $[iso / 1000.0, expotime * 1000, fnumber, speed, flength / 10.0, biasvalue]$ is used for normalization.

\subsection{Details of the Artist's Operations}\label{sec. operations}

\begin{figure}
\centering
\includegraphics[width=\textwidth]{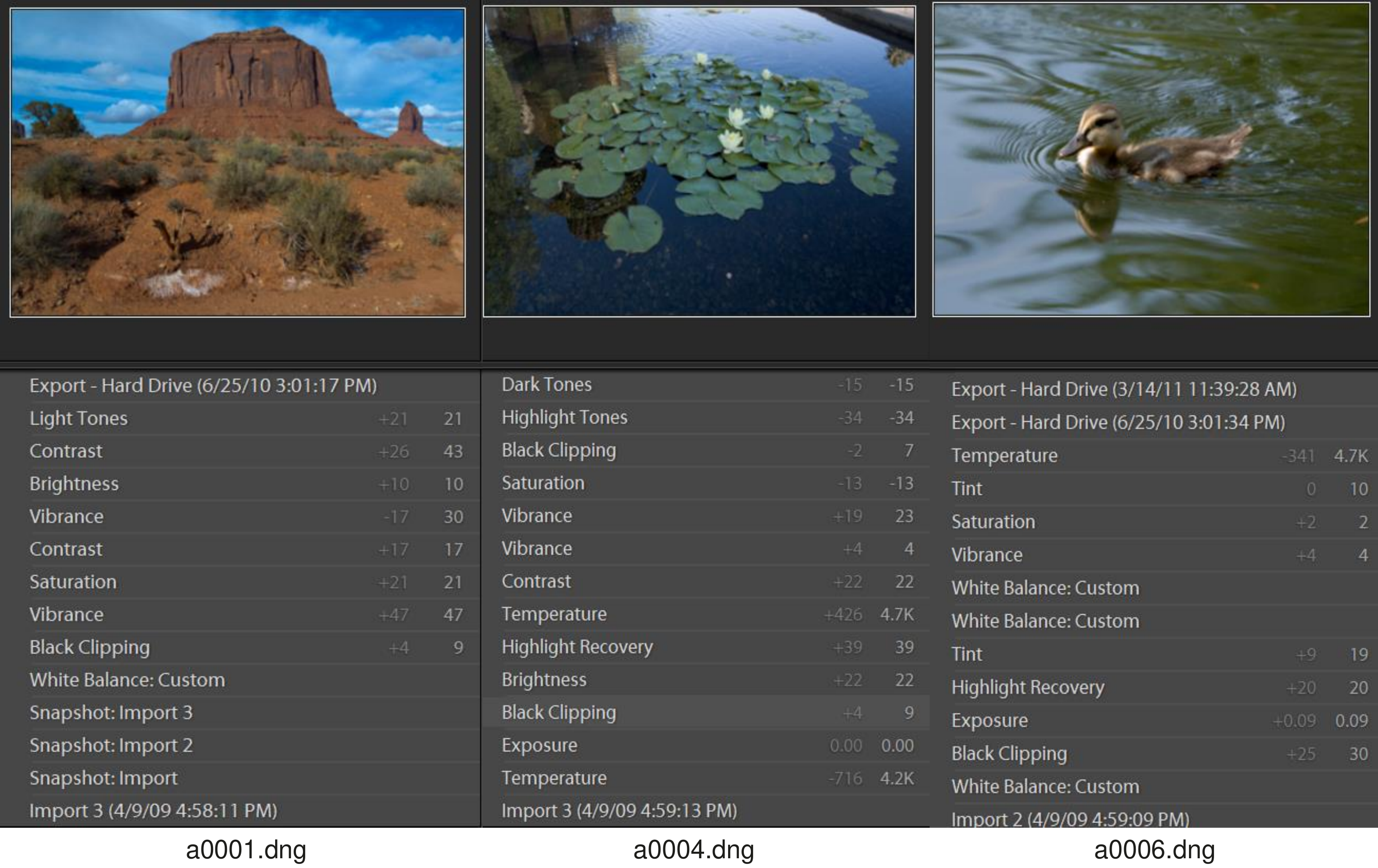}
\caption{The artist's retouching operations in a single image, the software is Lightroom.}
\vspace{-0.6cm} 
\label{fig:operations}
\end{figure}

A pie chart is used in the main paper to show the proportion of the artist's retouching operations, and the detailed operation data is in Table\ref{table:operations}. The artist's operation is obtained from the lightroom of the original dataset, and the operation steps of a single image in lightroom are shown in Figure \ref{fig:operations}.

\setlength{\tabcolsep}{4pt}
\begin{table}
\begin{center}
\caption{The specific statistics of the artist's operations, the total number of operations in the last two steps is less than 500 because the operation steps of some pictures are less than 5 steps. Operations not involved are marked as (“/”).}
\label{table:operations}
\begin{tabular}{ccccccc}
\hline\noalign{\smallskip}
Operations & The first step & The second step &The third step& The fourth step & The fifth step \\
\noalign{\smallskip}
\hline
\noalign{\smallskip}
Black Clipping &5 &26 &174&75& 60  \\
Brightness &/  &1 &/ &6&6 \\
Contrast &4&5&42  & 53 & 46\\
Dark Tones &/ &1&2&7 &15 \\
Exposure&447  & 41 &30 & 11 & 15 \\
Fill light&/& 1 & 6 &13&14\\
Highlight Recovery &17 &391 &41 &27& 15\\
Light Tones &/&/& / &2&13\\
Midtone Split &/&/ &/ &/&1\\
Highlight Tones &/&/  &/&/&4\\
Saturation &/ &/ & 1&9&23  \\
Shadow Tones &/ &/ &1&1&5  \\
Temperature &2 &2 & 5&24& 42  \\
Tint& / &1 &4 &5&7  \\
Vibrance &/ &8 & 133&173& 108  \\
White Balance &25 &23 & 60&89& 86  \\
\hline
Total &500 &500 & 500&495& 460  \\
\hline
\end{tabular}
\end{center}
\end{table}
\setlength{\tabcolsep}{1.4pt}

\subsection{Hue Palette Loss Implementation Details}\label{sec. hueloss}

\begin{figure}
\centering
\includegraphics[width=\textwidth]{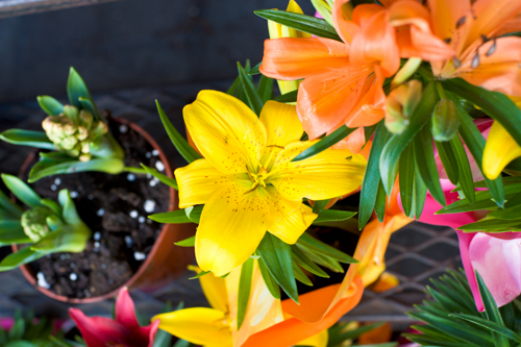}
\caption{The 959th image in the MIT Adobe-fiveK dataset}
\label{fig:a959}
\end{figure}

\begin{figure}
\centering
\includegraphics[width=\textwidth]{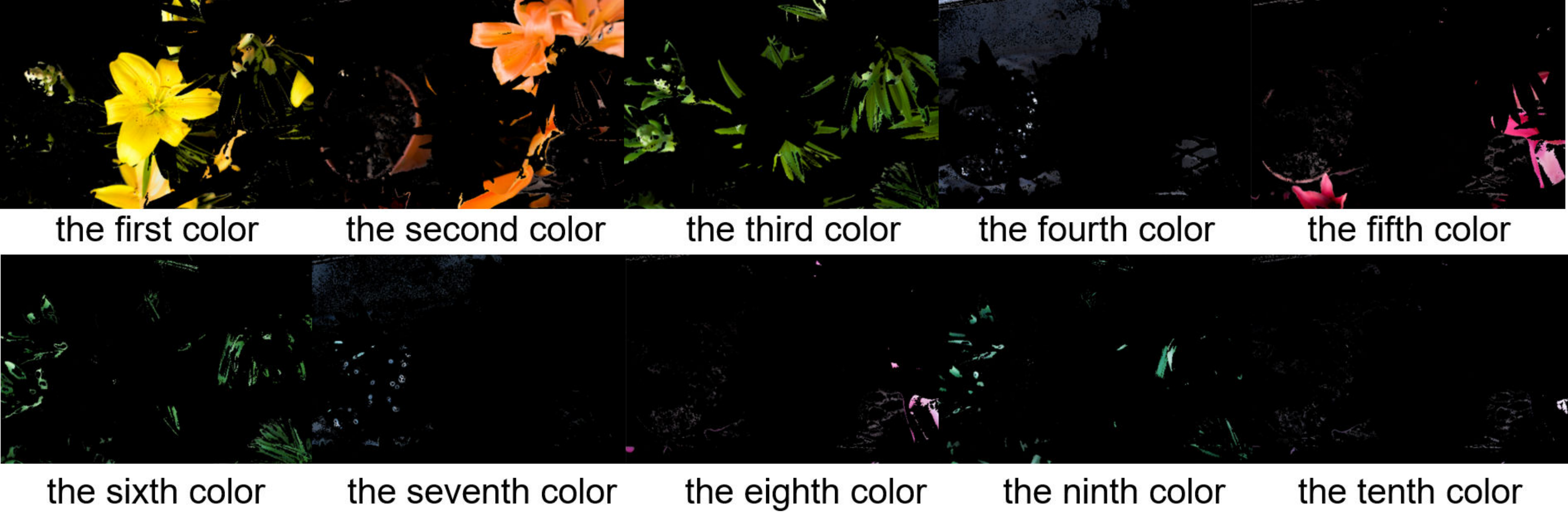}
\caption{Ten Colors Extracted}
\label{fig:10colors}
\end{figure}

The hue palette loss designed needs to specify the number of divided histogram intervals, that is, the number of classes of colors. The influence of the number of different intervals on the final LCCNet results is shown in the table. In this histogram, too wide of bin width can cause color confusion, while too narrow will cause similar colors to be divided into different ranges. We divide the H channels in the range of [0, 360] into 10 equal bins by experiments. For a single image \ref{fig:a959}, the ten colors after extraction are shown in Figure \ref{fig:10colors}.

\setlength{\tabcolsep}{4pt}
\begin{table}[htbp]
\caption{The Effect of the Number of Colors on Hue Loss(480P).} 
\begin{center}
\scalebox{0.9}{
\label{table:histogram}
\begin{tabular}{cccccccccc}
\hline\noalign{\smallskip}
number of intervals & 5 & 6 & 8 & 9& 10& 12& 20& 30\\
\noalign{\smallskip}
\hline
\noalign{\smallskip}
PSNR  & 25.74dB & 25.69dB & 25.68dB & 25.77dB & 25.81dB & 25.68dB& 25.70dB & 25.68dB \\
\hline
\end{tabular}}
\end{center}
\end{table}
\setlength{\tabcolsep}{1.4pt}

\subsection{Qualitative comparison of LCCNet Modules}\label{sec. ablations}

\begin{figure}
\centering
\includegraphics[width=\textwidth]{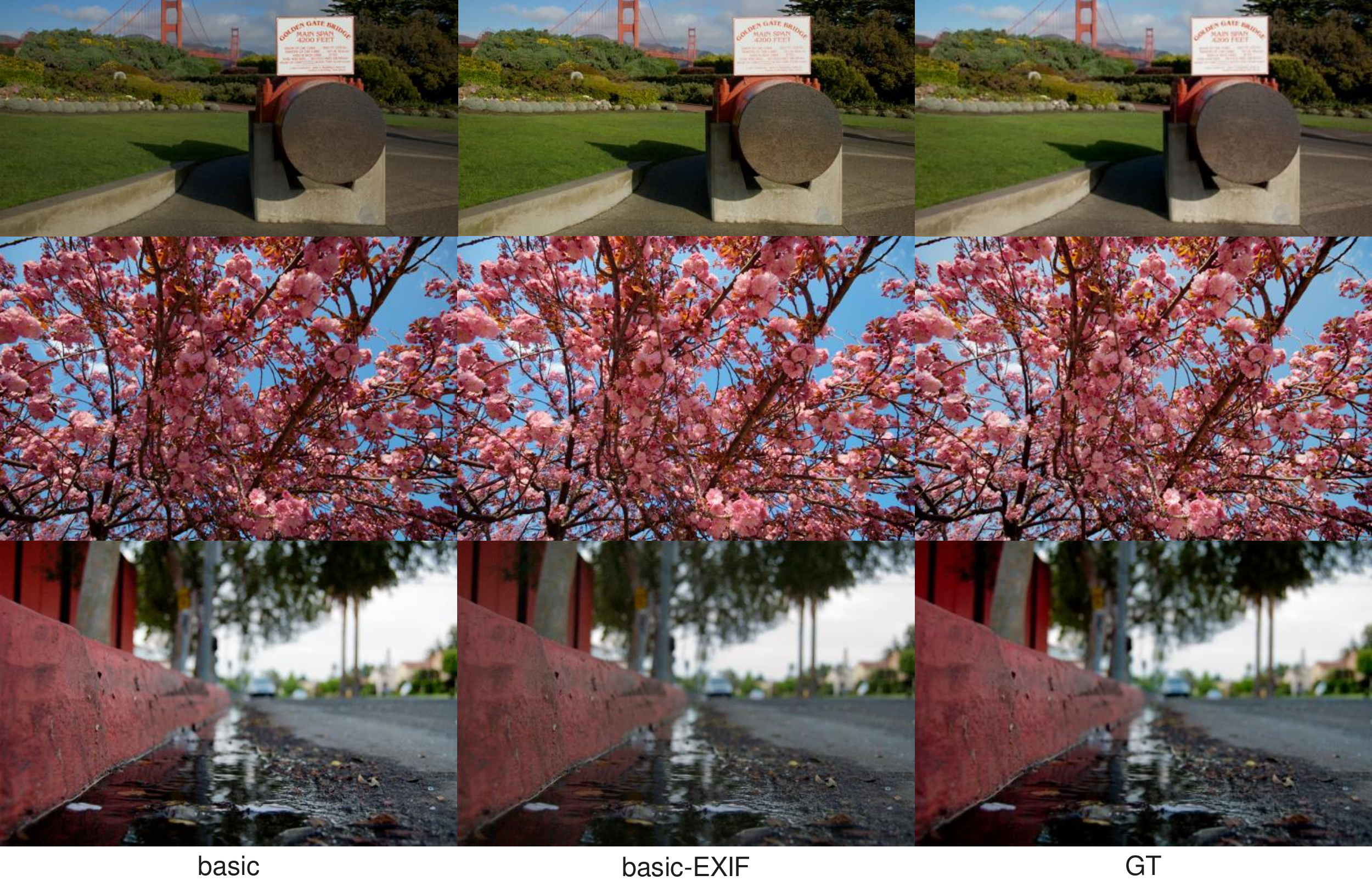}
\caption{Qualitative Comparison Between Basic Model and Basic Model-EXIF.}
\label{fig:ablation-EXIF}
\end{figure}

\begin{figure}
\centering
\includegraphics[width=\textwidth]{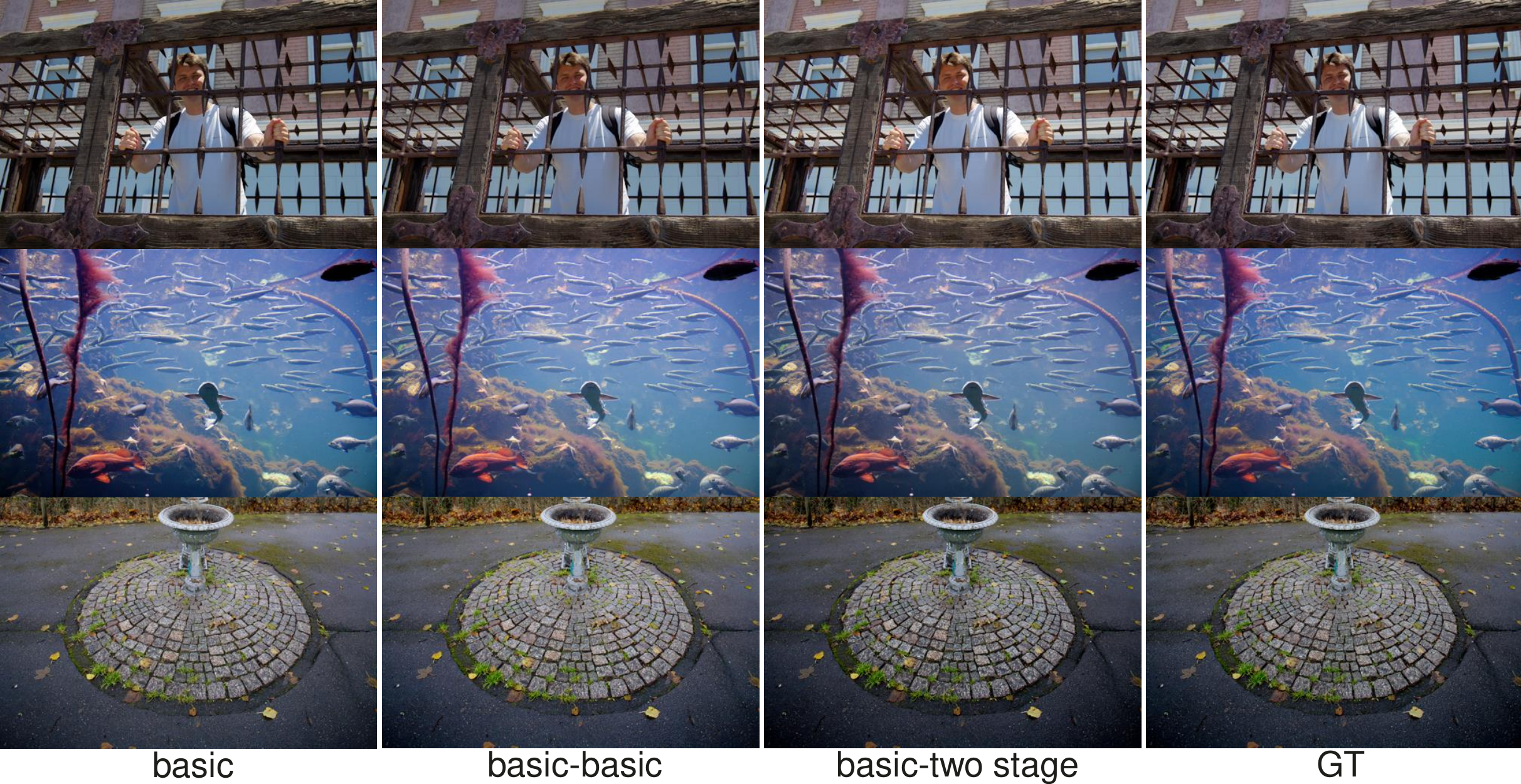}
\caption{Qualitative Comparison Between Basic Model and Basic Model-two stage.}
\label{fig:ablation-twostage}
\end{figure}

\begin{figure}
\centering
\includegraphics[width=\textwidth]{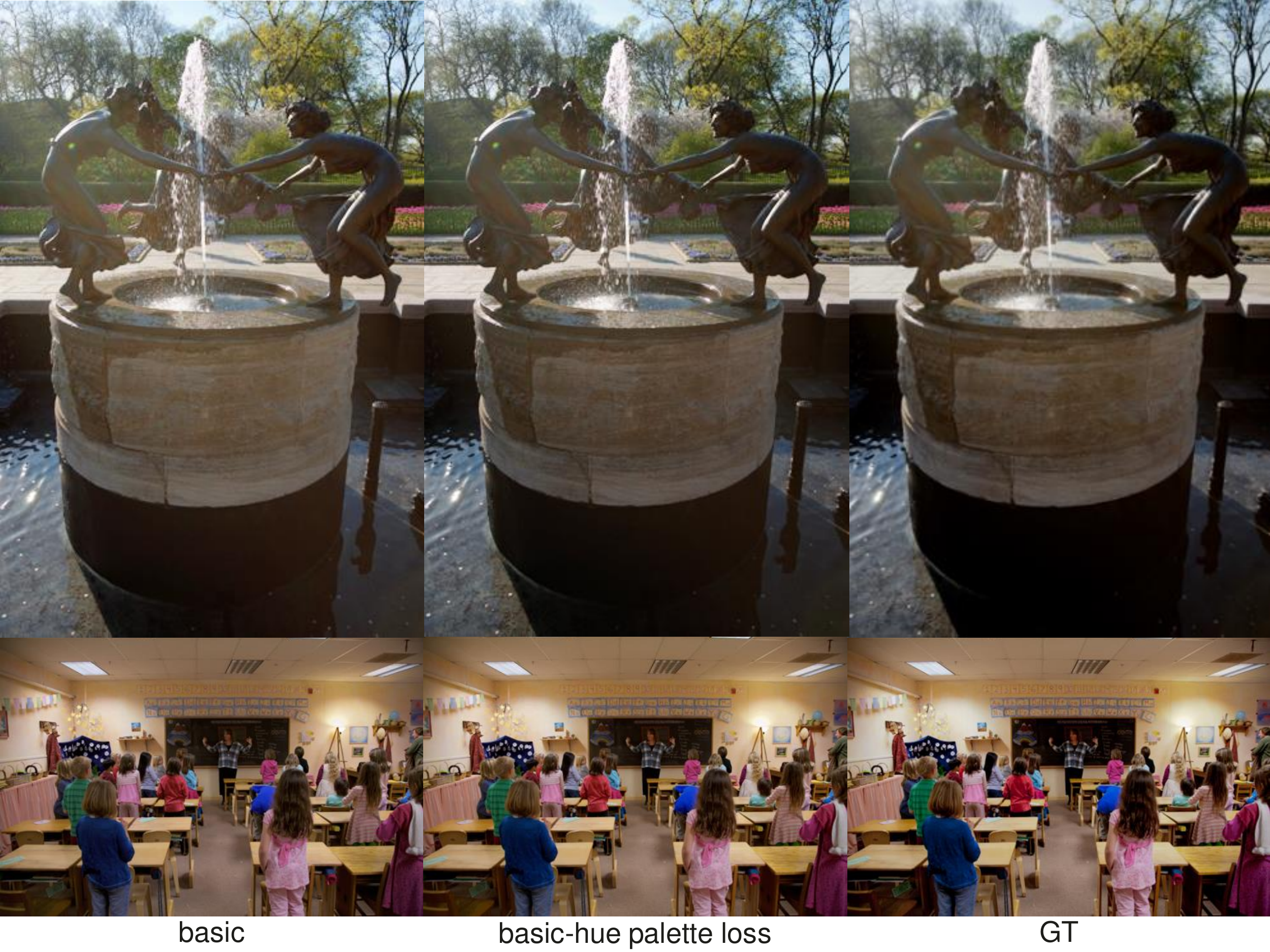}
\caption{Qualitative Comparison Between Basic Model and Basic Model-hue palette loss.}
\label{fig:ablation-loss}
\end{figure}

We conducted experiments on the effects of each module in LCCNet. Figure \ref{fig:ablation-EXIF} shows the comparison results of basic model and basic model-EXIF, 
the addition of EXIF makes the image closer to GT in chrominance. Figure \ref{fig:ablation-twostage} shows the comparison results of basic model and basic model-two stage, in addition, the simple cascaded form basic-basic without $gainmap$ is also compared. The structure of the two stage makes the luminance more realistic. Figure \ref{fig:ablation-loss} shows the comparison results of basic model and basic model-hue palette loss. It can be seen from the clothes that the hue loss that added make the color more saturated.

\subsection{Additional Comparison with Contemporaneous Methods}\label{sec. comparsions}

\begin{figure}[H]
\centering
\includegraphics[width=\textwidth]{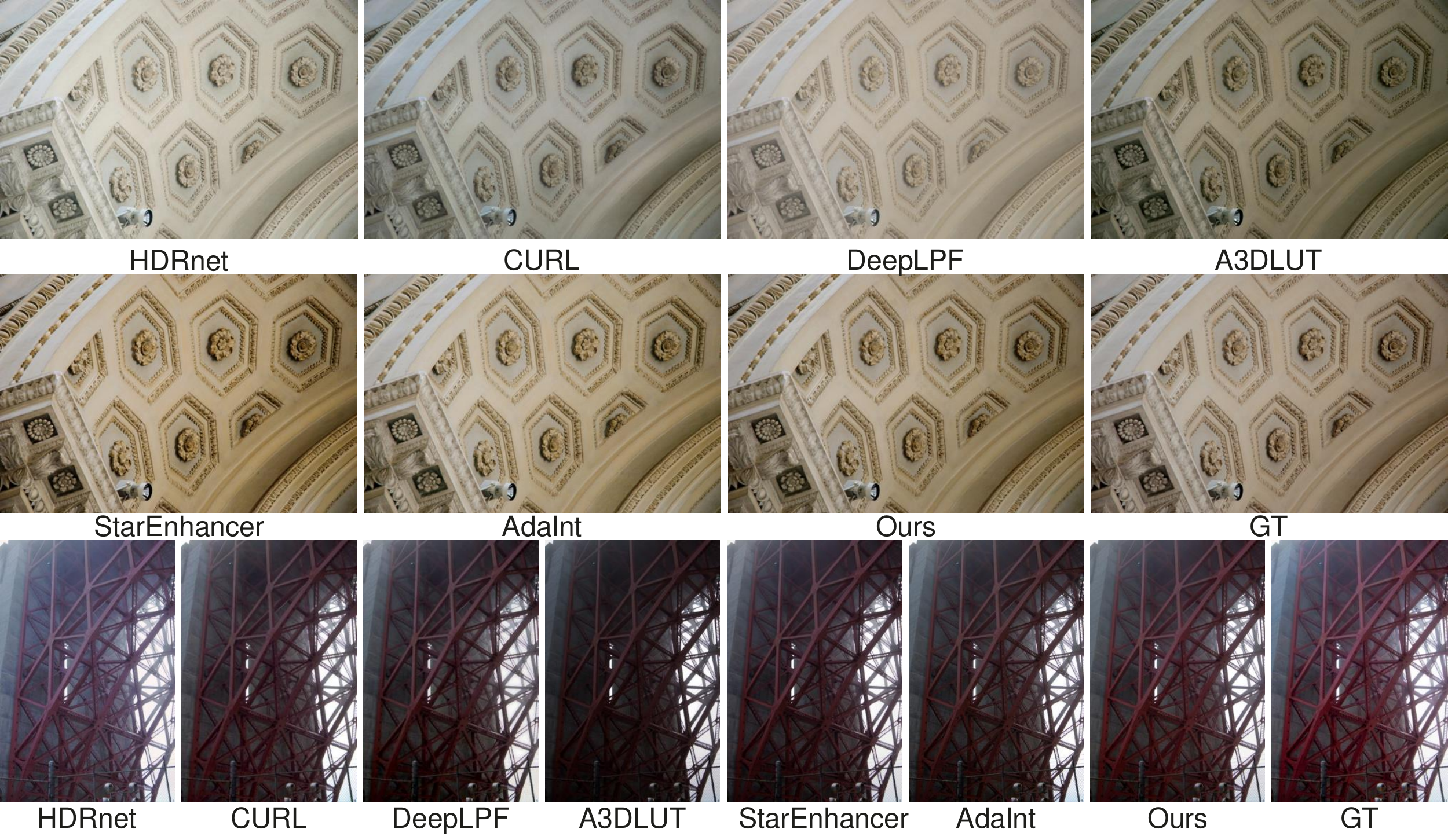}
\caption{Qualitative Comparison with Contemporaneous Methods on MIT-Adobe fiveK.}
\label{fig:comparsion1}
\end{figure}

\begin{figure}[H]
\centering
\includegraphics[width=\textwidth]{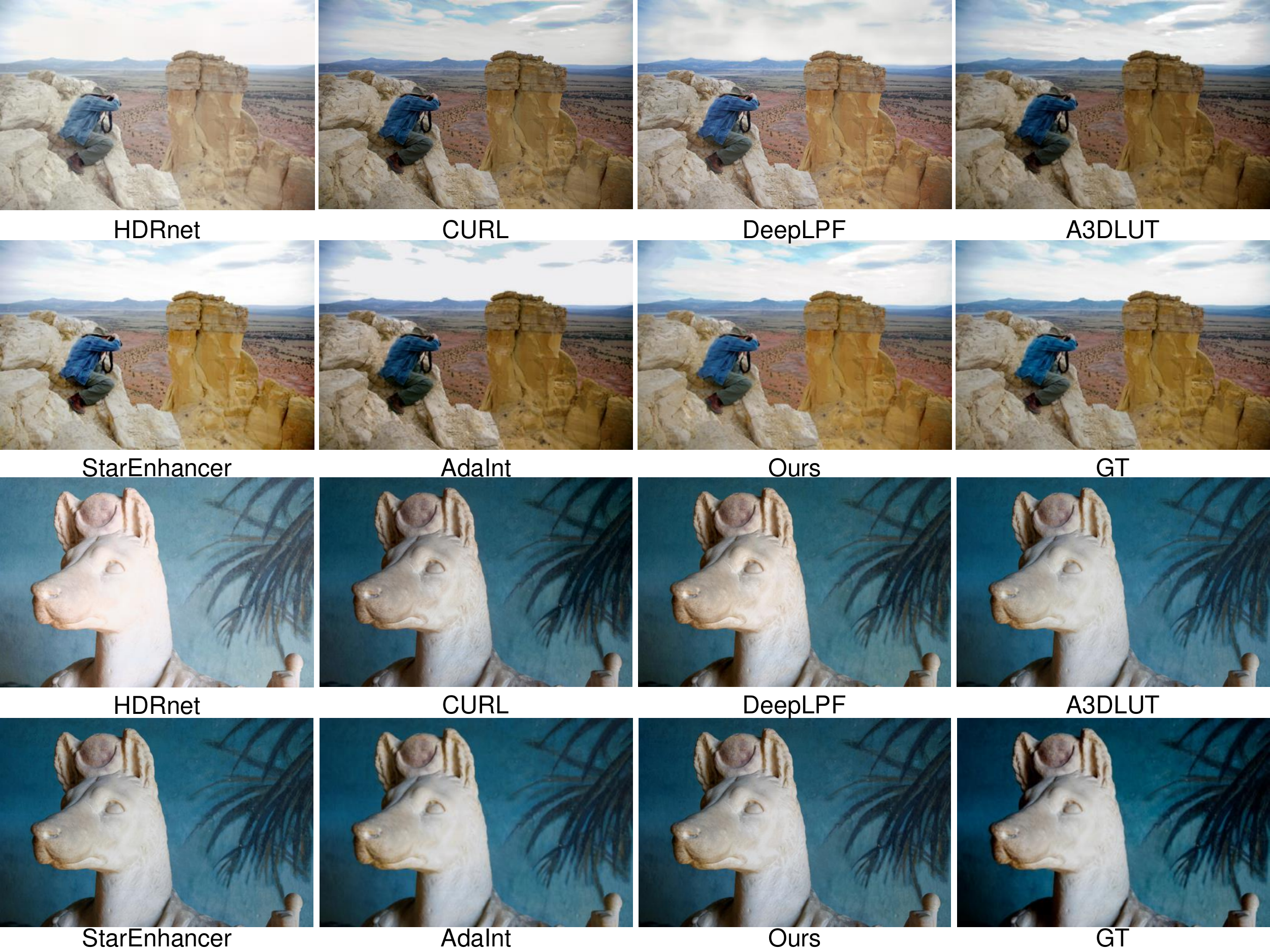}
\caption{Qualitative Comparison with Contemporaneous Methods on MIT-Adobe fiveK.}
\vspace{-0.6cm} 
\label{fig:comparsion2}
\end{figure}

\begin{figure}[H]
\centering
\vspace{-0.3cm} 
\includegraphics[width=\textwidth]{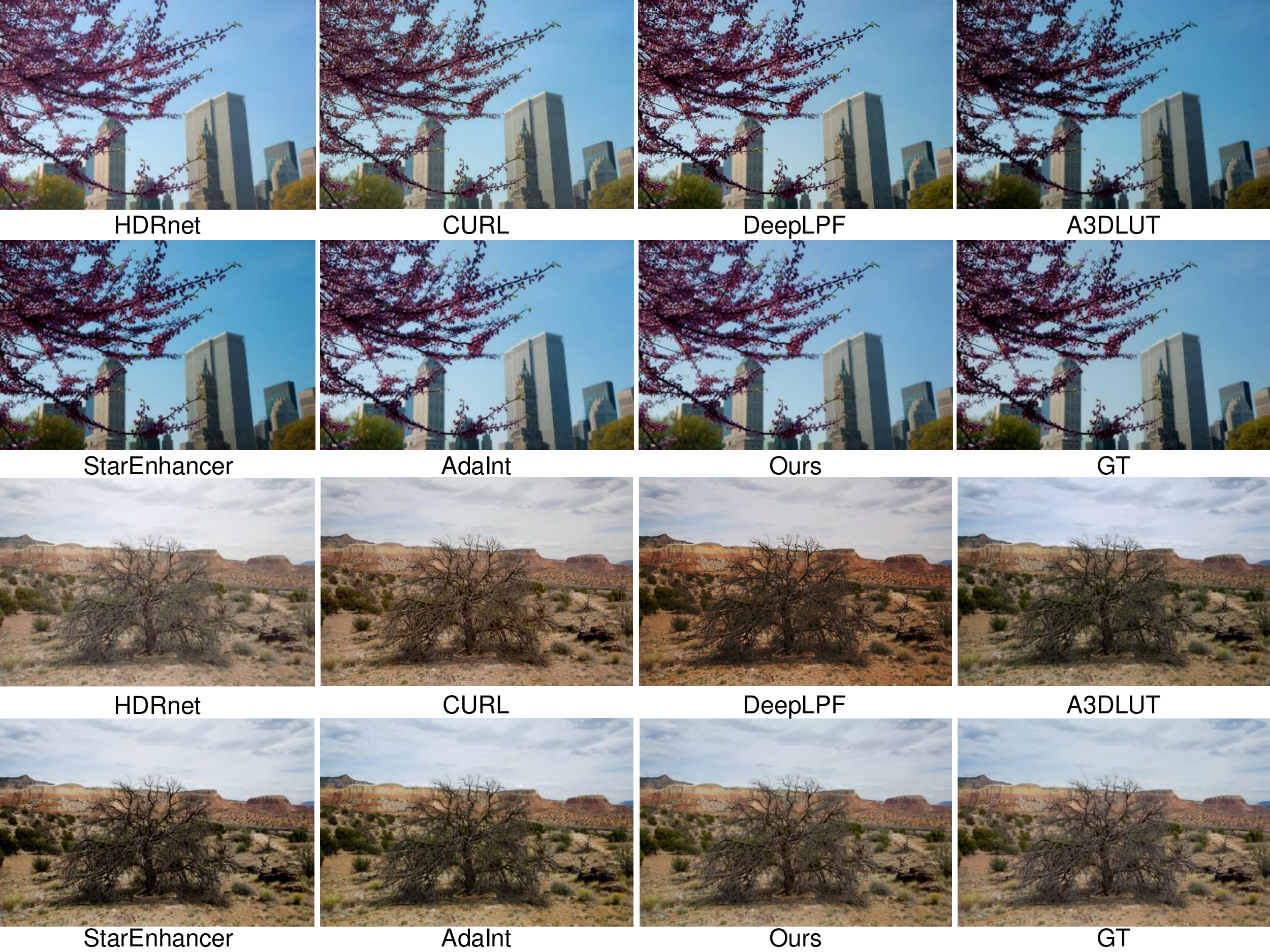}
\vspace{-0.3cm} 
\caption{Qualitative Comparison with Contemporaneous Methods on MIT-Adobe fiveK.}
\label{fig:comparsion3}
\end{figure}

An additional comparison methods with the contemporaneous method, including HDRNet~\cite{gharbi2017hdrnet}, CURL~\cite{moran2021CURL}, DeepLPF~\cite{moran2020deeplpf}, A3D-LUT~\cite{zeng20203Dlut}, StarEnhancer~\cite{Song2021starenhancer}, AdaInt~\cite{yang2022AdaInt}. The results can be seen in Figure \ref{fig:comparsion1}, Figure \ref{fig:comparsion2} and Figure \ref{fig:comparsion3}. In Figure \ref{fig:comparsion1}, our method 
closer to GT in chrominance, and the second scene has the closest brightness performance to the reality. In Figure \ref{fig:comparsion2}, our sky color is the most realistic, the foreground and background of the stone statue are close to GT. In Figure \ref{fig:comparsion3} both the flowers in the first scene and the grass in the second are more realistic in chrominance, at the same time, the deviation between the brightness and the real scene is small.


{\small

\bibliographystyle{ieee}
}








\end{document}